\documentclass[letterpaper,journal]{IEEEtran}
\usepackage{amsmath,amsfonts}
\usepackage{algorithmic}
\usepackage{algorithm}
\usepackage{array}
\usepackage[caption=false,font=normalsize,labelfont=sf,textfont=sf]{subfig}
\usepackage{textcomp}
\usepackage{stfloats}
\usepackage{url}
\usepackage{verbatim}
\usepackage{graphicx}
\usepackage{cite}
\usepackage[utf8]{inputenc}
\usepackage[T1]{fontenc}
\usepackage{hyperref}
\usepackage{booktabs}
\usepackage{nicefrac}
\usepackage{microtype}
\usepackage{xcolor}
\usepackage{siunitx}
\usepackage{multirow}
\usepackage[normalem]{ulem}
\usepackage{makecell}
\usepackage{natbib}
\usepackage{tabularx}
\usepackage{caption}
\begin{document}

\title{Reprojection-Guided 3D Gaussian Splatting Diffusion for Weakly Supervised Single‑Image Normal Estimation}

\author{Yanxing Liang, Yinghui Wang*, Min Wu, Wei Li*~\IEEEmembership{Senior Member, IEEE}, Tao Yan~\IEEEmembership{Senior Member, IEEE}, Jiaxing Shen
\thanks{This work was supported by in part of the ``National Key Research and Development Program of China (No.2023YFC3805901)'', the ``National Natural Science Foundation of China (No.62172190)'', in part of the ``Taihu Talent-Innovative Leading Talent Plan Team of Wuxi City (Certificate Date: 20241220(8))''.}
\thanks{Yanxing Liang, Yinghui Wang, Min Wu, Wei Li and Tao Yan are with the School of Artificial Intelligence and Computer Science, Jiangnan University, Wuxi, China.}
\thanks{Jiaxing Shen is with the School of Data Science, Lingnan University, Hong Kong, China.}
\thanks{Corresponding authors: Yinghui Wang (e-mail: wangyh@jiangnan.edu.cn) and Wei Li (e-mail: cs$\_$weili@jiangnan.edu.cn).}}

\IEEEaftertitletext{\vspace{-1\baselineskip}}

\markboth{Journal of \LaTeX\ Class Files,~Vol.~xx, No.~x, xx~xx}%
{Yanxing Liang \MakeLowercase{\textit{et al.}}: Reprojection-Guided 3D Gaussian Splatting Diffusion for Weakly Supervised Single‑Image Normal Estimation}

\maketitle

\begin{abstract}
Estimating object surface normals from a single RGB image is inherently ill-posed. Existing weakly supervised methods, whether discriminative models that learn mappings via indirect proxy supervision or generative models that reconstruct geometry based on implicit shape priors, fail to establish a physically consistent correspondence between 2D image observations and 3D geometry. Consequently, the normals they produce are geometrically inconsistent with the underlying surface, yielding physically implausible results.
To address this issue, we propose \textbf{CLONE}, a \textbf{C}ontinuous \textbf{L}atent \textbf{O}ptimization framework for \textbf{N}ormal \textbf{E}stimation via 3D Gaussian splatting. The core idea is to construct an image-geometry-image consistency strategy that unifies explicit geometric representation with differentiable rendering, thereby enabling weakly supervised learning without normal ground truth.
Specifically, CLONE comprises four components. First, by introducing a differentiable light interaction model with a learnable modulation kernel, we perform a unified reparameterization of the 3DGS parameter space. Second, the conditional single-step deterministic refinement network integrates denoising architectures with differentiable reprojection constraints to refine the initial normals, thereby adaptively recovering the high-frequency details erased by the inherently smooth Gaussian primitives. Third, the cross-domain gating fusion mechanism adaptively combines the two complementary normal estimates, reconciling the geometrically consistent yet over-smooth 3DGS estimate with the detailed yet potentially geometry-inconsistent refinement. Finally, all components are jointly optimized under a unified photometric reprojection objective with geometric consistency regularizations in a fully differentiable pathway, achieving an end-to-end optimization closed loop without relying on external normal labels.
Trained only on images paired with their corresponding 3D models, without any pixel-level normal annotations, our method achieves state-of-the-art performance among weakly supervised methods and delivers accuracy competitive with fully supervised baselines.
By eliminating the dependence on pixel-level normal annotations, the proposed framework provides a more label-efficient alternative for high-quality single-image normal estimation.
\end{abstract}

\begin{IEEEkeywords}
Single-Object Normal Estimation, 3D Gaussian Splatting, Conditional Diffusion, End-to-End Differentiable Optimization.
\end{IEEEkeywords}

\section{Introduction}\label{sec1}

Estimating object surface normals from a single RGB image is a fundamental task in computer vision and computer graphics. Recovering the 3D orientation of each visible surface point from a single 2D observation without multi-view cues leads to multi-solution ambiguity.
Most state-of-the-art methods adopt the fully supervised paradigm and rely on large-scale pixel-level normal ground truth for training. High-precision normal ground truth can only be obtained via 3D scanning or fine-grained datasets to resolve multi-view ambiguity. Collecting diverse normal labels restricts the scalability of these approaches to novel object classes and diverse imaging scenes.
Motivated by this challenge, weakly supervised normal estimation methods that dispense with pixel-wise normal ground truth have become a research avenue with prominent practical and theoretical value. However, these methods inherit the same multi-solution ambiguity, amplified by the absence of normal ground truth.

Existing weakly supervised methods pursue this goal through two technical routes.
Discriminative methods learn a direct image-to-normal mapping under proxy supervision signals~\cite{Ladicky2014Discriminatively, Wang2015Designing, Eigen2015Predicting, Yin2019VirtualNormal}. The mapping is trained to produce normals that are consistent with the proxy signal, with the optimization conducted in the normal space. The constraint this mechanism brings to the ill-posed problem arises from the learned statistical relationship between image features and normals encoded in the network weights.
Generative methods, by contrast, infer normals from a learned prior distribution over 3D shapes~\cite{Lin2023Magic3D, Melas-Kyriazi2023RealFusion, Liu2024SyncDreamer}. The inference is mediated by the prior's latent space, with the image providing conditional guidance. The constraint this mechanism brings arises from the prior's statistical regularities, which ensure the output is a plausible geometry.
These two constraints address different aspects of the ill-posed problem. The discriminative mapping anchors the prediction to the image content, ensuring the output is responsive to the observed input. The generative prior anchors the inference to plausible geometry, ensuring the output is structurally meaningful. Single-view normal estimation fundamentally requires both properties: the normals must be consistent with the image observation and must correspond to a physically plausible surface. The two constraints are therefore complementary: each supplies what the other does not address, and the problem can only be fully constrained when both are brought to bear.

\begin{figure*}[!t]
    \centering
    \includegraphics[width=0.8\linewidth]{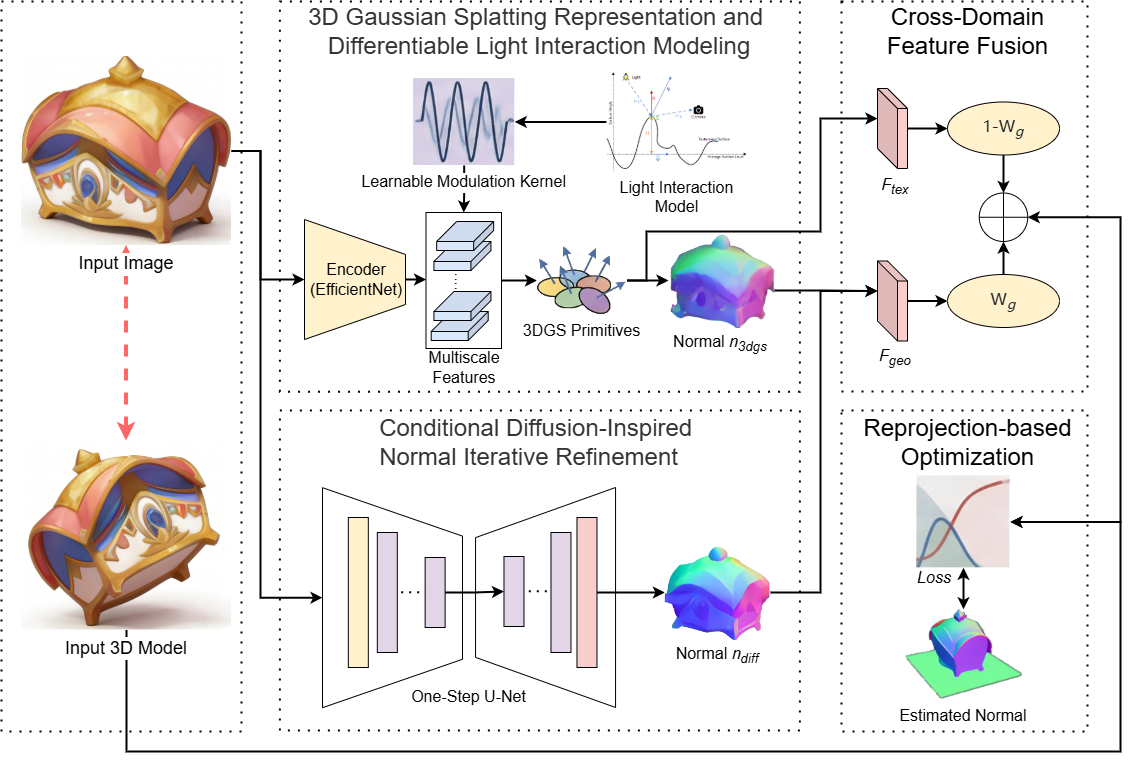}
    \caption{Overview of the CLONE pipeline. Given a single RGB image, the framework predicts 3DGS parameters, derives geometric normals, refines them via a single-step deterministic refinement network and adaptively fuses both estimates. The entire pipeline is supervised end-to-end by a photometric reprojection loss.}
    \vspace{-16pt}
    \label{fig-pipeline}
\end{figure*}

To combine the two routes, we propose CLONE, a \textbf{C}ontinuous \textbf{L}atent \textbf{O}ptimization framework for \textbf{N}ormal \textbf{E}stimation via 3D Gaussian splatting (Fig.~\ref{fig-pipeline}). The framework is structured around a shared 3D Gaussian representation that serves as the bridge between the 2D image domain and the 3D geometry domain.
Specifically, a differentiable reparameterization of the 3DGS parameter space establishes a stable analytical mapping from geometric primitives to surface normals, and a physically-based illumination model with a learnable modulation kernel provides a fully differentiable forward rendering pathway. Through this pathway, the photometric discrepancy between the rendered image and the input observation propagates gradients directly to the 3D geometry, making the image-derived constraint a signal that continuously drives the 3D representation toward consistency with the observation.
A conditional single-step deterministic refinement network, operating as a noise-conditioned residual operator on the initial geometric normals, recovers fine geometric details by leveraging the inductive bias of denoising architectures while remaining fully differentiable and compatible with the same photometric reprojection constraint.
The outputs of the explicit geometric representation and the refinement network reside in different spaces, one physically interpretable but smooth and the other detail-rich, and a cross-domain gating fusion mechanism adaptively reconciles the two, weighting each estimate according to local surface complexity and imposing multi-view reprojection consistency to ensure geometric plausibility.
All components are jointly optimized under a unified photometric reprojection objective with geometric consistency regularizations, forming an end-to-end differentiable loop in which the image-derived constraint informs the geometry and the geometric constraint is verified against the image, without relying on external normal labels.

The main contributions of our method are summarized as follows:
\begin{itemize}
    \item A unified framework that combines feedforward geometric priors with differentiable photometric verification through an image-geometry-image consistency loop, providing a self-supervising mechanism for weakly supervised normal estimation that corrects both proxy inaccuracy and prior deviation without external normal labels.
    \item A high-fidelity differentiable illumination rendering pipeline with a learnable modulation kernel that turns the photometric loss into an internal supervision signal under weak supervision.
    \item A single-step deterministic refinement network that recovers high-frequency detail while preserving end-to-end differentiability, without external label data.
    \item A cross-domain gating fusion mechanism with multi-view reprojection consistency and implicit geometric regularization that reconciles geometric consistency with recovered detail, without relying on external normal labels.
\end{itemize}

Extensive experiments on multiple object-level benchmarks show that CLONE achieves state-of-the-art performance among weakly supervised methods and accuracy competitive with fully supervised baselines.

\section{Related Work}\label{sec2}

\textbf{Convolutional Neural Networks and Transformers.}
Early works formulate normal estimation as a dense prediction problem, using CNNs or Transformer-based architectures to learn a direct mapping from image pixels to surface normals~\cite{Ladicky2014Discriminatively, Wang2015Designing, Eigen2015Predicting, Yin2019VirtualNormal}. These methods typically adopt multi-scale feature aggregation and incorporate geometric priors to improve prediction accuracy. For instance, Wang \textit{et al.}~\cite{Wang2015Designing} jointly model normal estimation and indoor layout prediction, while Yin \textit{et al.}~\cite{Yin2019VirtualNormal} introduce virtual normal constraints to enhance robustness.
Although these methods perform well under supervised settings, they rely heavily on dense normal labels, which are expensive and limited in diversity. More importantly, they operate entirely in the 2D domain and treat normal estimation as a pixel-wise regression problem without explicitly modeling the underlying 3D geometry. As a result, their optimization process mainly relies on supervision to implicitly constrain the solution space, lacking an intrinsic geometry-driven optimization mechanism, which limits their generalization under unseen illumination or viewpoint changes.

\textbf{Generative Models.}
Recent advances in generative modeling, especially diffusion-based methods, provide a new perspective by formulating normal estimation as an inverse generative problem. These methods infer plausible 3D structures from a single image by leveraging strong learned priors. Representative works such as Magic3D~\cite{Lin2023Magic3D}, RealFusion~\cite{Melas-Kyriazi2023RealFusion}, SyncDreamer~\cite{Liu2024SyncDreamer} and Zero123~\cite{Liu2023Zero123} demonstrate strong capabilities in single-view 3D reconstruction, with subsequent extensions further improving geometric fidelity.
Several parallel lines of work have addressed the gradient truncation problem in multi-step diffusion models. Consistency Models~\cite{Song2023Consistency} learn to map any point on the diffusion trajectory to the trajectory's endpoint in a single step. DDIM~\cite{Song2021DDIM} enables deterministic single-step inference through a non-Markovian forward process. Score Distillation Sampling (SDS)~\cite{Poole2023DreamFusion} bypasses backpropagation through the diffusion chain by using the score function as a guidance signal.
Despite their strong generative performance, these methods are predominantly driven by 2D priors and impose limited explicit constraints on 3D geometric consistency. Due to the absence of explicit geometric variables, they often produce ambiguous or inconsistent surface orientations. Their optimization relies on implicit distribution matching or iterative sampling processes, making it difficult to establish a stable mechanism for propagating geometric constraints from image observations, thereby limiting their effectiveness in optimization-driven geometric inference tasks.

\textbf{Explicit Geometric Modeling Based on 3DGS.}
Parallel to generative methods, recent works explore explicit 3D representations, particularly 3DGS, for geometric modeling and reconstruction. Compared with alternative 3D representations, 3DGS offers unique advantages. Depth maps only encode 2.5D surface information and lack full volumetric reasoning, whereas 3DGS provides a complete 3D scene parameterization with analytical gradient pathways. Point clouds are discrete and lack surface continuity, whereas Gaussian primitives intrinsically define a continuous density field from which surface normals can be derived via covariance eigen-decomposition. Methods such as SuGaR~\cite{Guedon2024SuGaR}, Normal-GS~\cite{Wei2024NormalGS} and GeoSplatting~\cite{Ye2025GeoSplatting} demonstrate that 3DGS provides an efficient and differentiable representation that enables gradient-based optimization via differentiable rasterization.
However, existing 3DGS-based methods are primarily designed for multi-view reconstruction or rely on additional supervision. In single-view settings, they lack sufficient constraints to uniquely determine the underlying geometry. More importantly, although 3DGS provides a differentiable forward rendering process, existing methods do not explicitly formulate normal estimation as a joint optimization problem and thus fail to fully exploit image-level errors as constraints on geometric variables. As a result, the potential of 3DGS for monocular geometric reasoning remains underexplored.

\section{Methodology}\label{sec3}

\subsection{Problem Formulation}\label{sec31}

Existing 3DGS methods are designed for multi-view dense reconstruction~\cite{Guedon2024SuGaR, Wei2024NormalGS, Ye2025GeoSplatting} and face four obstacles when adapted to single-view weak supervision. Standard parameterization lacks a stable analytical mapping from covariances to normals without normal ground truth. Native rendering pipelines sacrifice gradient accuracy for rasterization efficiency, providing insufficient fidelity for normal-level optimization. The intrinsic smoothness of Gaussian primitives erases high-frequency details, yet mainstream detail enhancement depends on normal ground-truth supervision. Naively fusing normals from geometric priors with those from detail enhancement breaks global consistency without ground-truth constraints on fusion boundaries.

Specifically, standard 3DGS parameterization targets color reconstruction as its primary objective and provides no stable analytical mapping between geometric covariances and surface normals. We address this by performing a unified reparameterization of the 3DGS parameter space: an encoder predicts per-pixel Gaussian parameters from a single image, and eigen-decomposition of the covariance matrices analytically yields surface normals. Together with rotation-scale decomposition and viewing direction constraints, this explicit geometry-to-normal mapping achieves sufficient numerical stability under single-view weak supervision and outputs physically reliable initial normals without any ground-truth supervision.

Native rendering pipelines are designed for rasterization efficiency rather than gradient accuracy and therefore cannot support normal-level geometric optimization. We replace this pipeline with an illumination rendering model that embeds a learnable oscillatory modulation kernel. The model implements a fully differentiable forward rendering process in which photometric error propagates gradients completely and accurately to the underlying 3D Gaussian parameters. Under this high-fidelity pipeline, photometric loss, combined with geometric regularizations, becomes a reliable direct supervision signal for geometric optimization, avoiding the gradient distortion inherent in native rasterization.

The intrinsic smoothness of Gaussian primitives erases high-frequency geometric details, yet mainstream detail enhancement schemes depend on normal ground-truth supervision that is unavailable in our setting. We resolve this tension through a conditional single-step deterministic refinement network. The network retains the inductive bias of the U-Net denoising architecture but reformulates its objective as a noise-conditioned residual refinement operator driven by photometric loss. Because the refinement is single-step rather than iterative, gradient truncation in multi-step diffusion is avoided entirely, and high-frequency geometric details are recovered without normal labels.

When normals from geometric priors and those from detail enhancement are naively fused, their fundamentally different representation spaces lead to physical distortion unless ground-truth constraints bind the fusion boundaries. We prevent this through a cross-domain gating fusion mechanism that adaptively weights globally reliable geometric normals against detail-rich refined normals. Multi-view reprojection consistency and implicit geometric regularizers further constrain the decomposition of albedo and shading, effectively alleviating the albedo-shading scale ambiguity. Meanwhile, directional regularization ensures the learnable principal directions align with the geometric normals, ensuring geometric plausibility of the output without direct normal supervision.

\subsection{3DGS Representation and Differentiable Light Interaction Modeling}\label{sec32}

3DGS explicitly models the target object with a set of 3D anisotropic Gaussian primitives. Compared with implicit function-based representations, such methods encode both geometry and appearance in an analytical form, offering advantages in interpretability and differentiable optimization. However, the parameters of standard 3DGS mainly serve rendering reconstruction, and there is no explicit and stable functional relationship between geometric parameters and surface normals, which limits its applicability in normal estimation. To address this, we perform a unified reparameterization of the 3DGS parameter space, preserving its explicit modeling advantages and introduce differentiable light interaction modeling to establish a continuous gradient path from 3D geometry to 2D observations.

We define all geometric quantities in a unified camera coordinate system, where the camera optical center is the origin, the optical axis is the positive $z$-axis, and the $x$- and $y$-axes in the image plane point to the horizontal and vertical directions, respectively. Given an input image $I(\boldsymbol{x})\in\mathbb{R}^{H\times W\times 3}$ with $\boldsymbol{x}=(u,v)$, we predict a set of 3D Gaussian primitives associated with image locations, while the final image formation follows a many-to-one splatting process in which multiple Gaussian primitives contribute to each pixel. Each Gaussian primitive is parameterized as:
\begin{equation}
    \label{eq311}
    G=\{\boldsymbol{\mu},\Sigma,\sigma,\boldsymbol{k}_D,\omega_g,\xi,\psi,\boldsymbol{d}\},
\end{equation}
where $\boldsymbol{\mu}\in\mathbb{R}^3$ is the Gaussian center, $\Sigma\in\mathbb{R}^{3\times 3}$ is the covariance matrix, $\sigma$ is the opacity parameter, $\boldsymbol{k}_D\in\mathbb{R}^3$ denotes the RGB diffuse albedo, $\omega_g$, $\xi$, $\psi$ control the frequency, spatial decay and phase shift of local light interaction, respectively, and $\boldsymbol{d}$ denotes a learnable local principal direction used to approximate the effect of normal variation on light modulation. During optimization, $\boldsymbol{d}$ is regularized to be consistent with the geometric normal via a small directional loss term (see Section~\ref{sec35}) to prevent degenerate solutions.

To ensure the symmetric positive definiteness of the covariance matrix during optimization and avoid numerical instability caused by direct optimization, we parameterize it using a rotation-scale decomposition:
\begin{equation}
    \label{eq312}
    \Sigma=RSR^\top,
\end{equation}
where $S=\mathrm{diag}(s_x^2,s_y^2,s_z^2)$ with $s_x,s_y,s_z>0$. The rotation matrix $R$ is mapped from a unit quaternion $\boldsymbol{q}\in\mathbb{S}^3$. Due to the sign equivalence of quaternions, where $\boldsymbol{q}$ and $-\boldsymbol{q}$ correspond to the same rotation, we impose a non-negativity constraint on its scalar component during training to avoid gradient discontinuities in backpropagation and select a unique representation in the parameter space.

Under this parameterization, each Gaussian primitive can be regarded as a continuous approximation of the local surface. Specifically, by performing eigen-decomposition on the covariance matrix:
\begin{equation}
    \label{eq313}
    \Sigma=U\Lambda U^\top,
\end{equation}
where $\Lambda=\mathrm{diag}(\lambda_1,\lambda_2,\lambda_3)$ and $\lambda_1\leq\lambda_2\leq\lambda_3$, the eigenvector $\boldsymbol{u}_{min}$ corresponding to the smallest eigenvalue gives the principal normal direction of the local surface. To eliminate sign ambiguity and ensure consistency in illumination computation, the normal direction is disambiguated by enforcing $\boldsymbol{n}_{3dgs}^\top \boldsymbol{v} > 0$, where $\boldsymbol{v}_i = \boldsymbol{\mu}_i / \|\boldsymbol{\mu}_i\|_2$ denotes the viewing direction from the camera optical center to the $i$-th Gaussian center in the camera coordinate frame. Unit normalization is applied to obtain the final geometric normal, where $\|\cdot\|_2$ denotes the Euclidean $\ell_2$ norm of a vector:
\begin{equation}
    \label{eq314}
    \boldsymbol{n}_{3dgs} = \frac{\boldsymbol{u}_{min}}{\|\boldsymbol{u}_{min}\|_2}.
\end{equation}

This process is fully built upon 3D geometric structures, avoiding uncertainties introduced by inferring normals directly from 2D feature spaces.

To connect the above geometric representation with image observations without normal labels, we introduce differentiable illumination constraints. Our illumination model is grounded in the Lambertian reflectance framework, which provides a physically principled baseline: radiance is proportional to the cosine between the surface normal and the light direction. However, real surfaces exhibit micro-geometric variations, such as roughness, facets and subsurface scattering that cause intensity perturbations beyond the smooth Lambertian prediction. The learnable modulation kernel $K(\boldsymbol{p})$ serves as a structure-enhancing multiplicative factor that captures these high-frequency residuals. Importantly, we design $K$ as an oscillatory function with a Gaussian envelope (Equation~\ref{eq317}) rather than a free-form learned weight map, for two reasons: the explicit oscillatory form prevents the kernel from degenerating into a texture-fitting function, since it cannot reproduce arbitrary image patterns, and the Gaussian envelope ensures the modulation is localized, preserving the global dominance of the Lambertian term.

Considering the diffuse imaging process under a single far-field light source, the radiance contribution of a single Gaussian primitive at pixel $\boldsymbol{x}$ is formulated as:
\begin{equation}
    \label{eq315}
    C_i(\boldsymbol{x}) = \boldsymbol{k}_{D,i} \odot \max(0,\boldsymbol{n}_{fuse,i}^\top\boldsymbol{l}) \cdot K_i(\boldsymbol{p}(\boldsymbol{x})),
\end{equation}
where ${\odot}$ denotes the element-wise multiplication between vectors, $\boldsymbol{l}$ denotes the light direction fixed to the camera optical axis for training, with robustness to varying illumination directions evaluated in Section~\ref{sec44}, $\boldsymbol{n}_{fuse,i}$ is the final fused normal of the $i$-th primitive, obtained by evaluating the pixel-level fusion $\boldsymbol{n}_{fuse}(\boldsymbol{x})$ at the projected 2D center of the $i$-th Gaussian, as detailed in Equation~\ref{eq333} of Section~\ref{sec34}, and $\boldsymbol{p}(\boldsymbol{x})$ denotes the 3D spatial point corresponding to pixel $\boldsymbol{x}$. The point $\boldsymbol{p}(\boldsymbol{x})$ is defined as the conditional expectation of the 3D position under the Gaussian density restricted to the camera ray:
\begin{equation}
    \label{eq316}
    \boldsymbol{p}(\boldsymbol{x}) = \frac{\int_{\mathbf{r}} \boldsymbol{p} \cdot \rho_G(\boldsymbol{p}) \, d\boldsymbol{p}}{\int_{\mathbf{r}} \rho_G(\boldsymbol{p}) \, d\boldsymbol{p}},
\end{equation}
where $\mathbf{r}$ denotes the camera ray corresponding to pixel $\boldsymbol{x}$ and $\rho_G(\boldsymbol{p})$ is the Gaussian density induced by the primitive along the ray. This provides a differentiable association between the 2D pixel and the 3D Gaussian. The final pixel radiance is obtained by blending contributions from all Gaussians, as detailed in Section~\ref{sec35}.

To obtain a numerically stable and learnable approximation, we model the local light field with an oscillatory kernel with a Gaussian envelope:
\begin{equation}
    \label{eq317}
    K(\boldsymbol{p})=\exp\left(-\xi\|\boldsymbol{p}-\boldsymbol{\mu}\|_2^2\right)\cos\left(\frac{2\pi}{\omega_g}\boldsymbol{d}^\top(\boldsymbol{p}-\boldsymbol{\mu})+\psi\right),
\end{equation}
where $\boldsymbol{p}-\boldsymbol{\mu}$ denotes the local coordinate relative to the Gaussian center, $\xi$ controls the spatial decay rate, $\omega_g$ controls the frequency distribution, $\boldsymbol{d}$ represents the local principal direction used to model the influence of normal variations on light modulation, and $\psi$ is the phase shift. This kernel serves as a first-order learnable approximation to local wave propagation rather than an exact physical diffraction simulation. It introduces tunable frequency response during optimization, enabling the model to capture high-frequency intensity perturbations caused by micro-geometry variations.

By embedding the above modulation kernel into the coupling term of albedo and normal, the observed image is sensitive to local geometric changes besides global illumination, providing additional geometric constraints in the unsupervised normal estimation framework. Combined with the explicit 3D Gaussian representation, this design establishes a continuous mapping from geometric parameters to image intensities in 3D space, allowing gradients to propagate stably from the pixel domain back to the Gaussian parameter space.

On this basis, our prediction network predicts the Gaussian parameter set for each spatial location conditioned on the input image. Predictions at different scales are fused at a unified resolution to form a multi-scale geometric feature representation. Since all geometric quantities are defined analytically and normals are directly derived from the covariance, the features have clear physical meanings and provide a stable and consistent geometric prior for subsequent refinement and differentiable reprojection optimization.

\subsection{Deterministic Normal Refinement via Single-Step Denoising}\label{sec33}

Although explicit modeling based on 3DGS can provide physically meaningful initial geometric estimation, it tends to exhibit significant over-smoothing in high-frequency details and complex structural regions due to the smooth nature of Gaussian primitives. On the other hand, recent diffusion model architectures have shown superior ability in capturing high-frequency structures and recovering details. However, their standard training paradigm relies on supervision signals from large-scale real data distributions, which is difficult to directly apply in single-image normal estimation without labels. To this end, we construct a conditional single-step deterministic refinement network. While preserving the structural inductive bias of U-Net denoising architectures, we reformulate its optimization objective as a noise-conditioned residual refinement operator driven by differentiable reprojection constraints defined later in Section~\ref{sec35}, thereby achieving high-quality normal refinement without ground-truth labels.

To formalize this process, we first introduce the basic formulation of continuous-time diffusion models. Let $\boldsymbol{n}_0$ denote the clean normal map. The forward diffusion process can be described by the variance-preserving stochastic differential equation:
\begin{equation}
    \label{eq321}
    d\boldsymbol{n}_t=-\frac{1}{2}\beta(t)\boldsymbol{n}_t dt+\sqrt{\beta(t)}dw_t,
\end{equation}
where $\beta(t)$ is a predefined noise schedule function and $w_t$ denotes a standard Wiener process. This process gradually evolves the original data distribution into an isotropic Gaussian distribution. Correspondingly, the reverse process can be expressed as a probability flow ordinary differential equation:
\begin{equation}
    \label{eq322}
    \frac{d\boldsymbol{n}_t}{dt}=-\frac{1}{2}\beta(t)\boldsymbol{n}_t-\beta(t)\nabla_{\boldsymbol{n}_t}\log p_t(\boldsymbol{n}_t),
\end{equation}
where $\nabla_{\boldsymbol{n}_t}\log p_t(\boldsymbol{n}_t)$ is the score function, which is usually approximated by a neural network.

In the unlabeled setting considered, the true distribution $p_0(\boldsymbol{n})$ is unavailable, invalidating standard diffusion training strategies. To address this issue, we reformulate the diffusion model from a probabilistic generator to a structural refinement operator. Specifically, the geometric normal $\boldsymbol{n}_{3dgs}$ obtained in Section~\ref{sec32} is used as the initialization rather than a direct supervision target. Given $\boldsymbol{n}_{3dgs}$, Gaussian noise is first injected at a random time step $t$. We discretize the continuous noise schedule $\beta(t)$ into a sequence $\{\beta_s\}_{s=1}^T$ and define the cumulative noise coefficient $\bar{\alpha}_t = \prod_{s=1}^t (1-\beta_s)$. Then the noisy normal is obtained as:
\begin{equation}
    \label{eq323}
    \boldsymbol{n}_t=\sqrt{\bar{\alpha}_t}\boldsymbol{n}_{3dgs}+\sqrt{1-\bar{\alpha}_t}\epsilon,\quad \epsilon\sim\mathcal{N}(0,\mathbf{I}),
\end{equation}
where $t$ is randomly sampled during training and is fixed to a predefined constant during inference. One-step denoising is then performed using a noise prediction network $\epsilon_\theta$ with time embedding to obtain the refined normal estimate:
\begin{equation}
    \label{eq324}
    \boldsymbol{n}_{\mathrm{diff}}=\frac{1}{\sqrt{\bar{\alpha}_t}}\left(\boldsymbol{n}_t-\sqrt{1-\bar{\alpha}_t}\epsilon_\theta(\boldsymbol{n}_t,t,c)\right),
\end{equation}
where $c = (F_{\mathrm{geo}}, F_{\mathrm{tex}})$ denotes the joint conditional features.

This one-step mapping can be regarded as a noise-conditioned residual refinement, where the network learns to correct the initial normal using local structure information. Since the whole process is converted into a deterministic forward propagation path conditioned on sampled noise, gradients of the differentiable reprojection loss can be directly back-propagated to the network parameters, enabling end-to-end optimization.

\subsection{Cross-Domain Feature Fusion Based on Gating Mechanism}\label{sec34}

The previous two sections model the normal estimation problem from the perspectives of explicit geometric modeling and generative refinement, respectively: the 3DGS-based representation provides physically meaningful initial normals and structural constraints in the 3D space, while the conditional deterministic module supplements high-frequency details in the 2D image domain. However, these two types of information have essential differences in expression form and statistical properties. The former relies on analytical geometric parameters and has a clear physical interpretation but limited expressive ability. The latter is derived from data-driven distribution learning, has strong detail modeling ability, but may introduce structures inconsistent with the true geometry. How to coordinate them in a unified framework to maintain both geometric consistency and detail expression ability is one of the core problems of this method.

From the perspective of model design, we introduce two complementary feature representations within the proposed framework. The geometric feature $F_{\mathrm{geo}}$ denotes the geometry-aware representation derived from the 3DGS-based prediction module, which encodes explicit 3D structural priors including Gaussian spatial properties, covariance-induced geometric cues and physical parameter dependencies. The texture feature $F_{\mathrm{tex}}$ denotes the appearance-aware intermediate feature extracted from the diffusion refinement network, which captures 2D image patterns, local edges and structural details that facilitate high-frequency detail recovery. Specifically, $F_{\mathrm{geo}}$ is the fused multi-scale feature map from the 3DGS encoder at $1/4$ input resolution, while $F_{\mathrm{tex}}$ is the intermediate feature map from the U-Net decoder at the same $1/4$ spatial resolution, and both are aligned via bilinear interpolation before fusion. Given the inherent domain discrepancy and semantic heterogeneity between 3D geometric priors and 2D appearance features, we introduce an explicit feature alignment and adaptive gating fusion mechanism to achieve effective cross-domain information interaction.

Specifically, we first map the geometric features to the same latent dimension as the texture features to establish a unified semantic space. On this basis, a joint attention-guided gating mechanism is introduced to adaptively weight the two types of features. Considering that geometric information dominates global structural reasoning while texture information contributes to local detail perception, we perform joint channel-spatial modulation on the input features. Channel-wise modulation learns to emphasize informative feature channels via global statistical aggregation, while spatial-wise modulation learns to locate informative regions via local context aggregation. The fused feature is then expressed as:
\begin{equation}
    \label{eq331}
    F_{\mathrm{fuse}}=W_g \odot F_{\mathrm{geo}} + W_t \odot F_{\mathrm{tex}},
\end{equation}
where ${\odot}$ denotes the element-wise multiplication between vectors, $W_g$ and $W_t$ are spatially and channel-wise adaptive weight maps. The weights are obtained via softmax normalization along the fusion branch dimension, i.e., $[W_g, W_t] = \mathrm{softmax}([M_{\mathrm{geo}}, M_{\mathrm{tex}}])$, where $M_{\mathrm{geo}}$ and $M_{\mathrm{tex}}$ denote the intermediate weight maps generated from the aligned features. This formulation ensures $W_g + W_t = \mathbf{1}$ in an element-wise manner, enabling complementary information fusion between geometry and texture.

After obtaining the unified conditional representation, we inject it into the diffusion network to perform conditional modulation in the high-level semantic space. Specifically, $F_b$ denotes the bottleneck feature of the diffusion U-Net, which aggregates global semantic information from the entire encoder pathway and serves as the query source for cross-attention.
Conditional guidance is then achieved through a cross-attention mechanism that aligns the unified conditional representation with the bottleneck feature. The attention operation is defined as:
\begin{equation}
    \label{eq332}
    \mathrm{Attn}(F_b,F_{\mathrm{fuse}})=\mathrm{softmax}\left(\frac{QK_{\mathrm{attn}}^\top}{\sqrt{d_k}}\right)V,
\end{equation}
where $Q = W_q \cdot \mathrm{Flatten}(F_b)$, $K_{\mathrm{attn}}=W_k \cdot \mathrm{Flatten}(F_{\mathrm{fuse}})$, $V=W_v \cdot \mathrm{Flatten}(F_{\mathrm{fuse}})$ and $d_k$ is the feature dimension of the attention keys. The attention output is integrated into the original bottleneck feature via a residual connection and layer normalization to maintain training stability.

Although the above fusion mechanism achieves collaborative modeling of geometry and texture in the feature domain, it is still necessary to explicitly combine the complementary normal estimates from the 3DGS geometric branch and the diffusion refinement branch. The geometric normal guarantees reliable large-scale structural consistency, while the refined normal restores rich high-frequency details. We therefore introduce a spatially adaptive gating mechanism to fuse the two complementary estimates. Let $\boldsymbol{n}_{3dgs}$ denote the geometric normal inferred from 3DGS and $\boldsymbol{n}_{\mathrm{diff}}$ denote the refined normal from the diffusion module. The final fused normal is defined as:
\begin{equation}
    \label{eq333}
    \boldsymbol{n}_{fuse}= g \odot \mathrm{sg}(\boldsymbol{n}_{3dgs}) + (1 - g) \odot \boldsymbol{n}_{\mathrm{diff}},
\end{equation}
where $\mathrm{sg}(\cdot)$ denotes the stop-gradient operation and the gating weight $g\in[0,1]$ is predicted by a lightweight convolutional network that takes the concatenation of the initial geometric normal and the input image as input. This fully differentiable formulation allows the photometric loss gradient to propagate through both branches simultaneously: $g$ adaptively weights the contribution of each normal estimate based on local surface complexity, learning to prioritize geometry ($g\to 1$) in flat regions and refinement ($g\to 0$) in texture-rich regions without disrupting the end-to-end gradient flow.

Through the above cross-domain fusion mechanism, geometric priors and generative refinement are no longer two independent stages, but form a tightly coupled relationship under a unified conditional modeling framework. This design realizes information complementarity at both feature and output levels, improving detail expression ability while maintaining physical consistency.

\subsection{Geometric Consistency Optimization Based on Differentiable Reprojection}\label{sec35}

In the aforementioned method, the 3DGS parameterization provides an explicit 3D geometric representation, the light interaction model defines the relationship between local normals and radiance, and the refinement module refines the normals. To achieve end-to-end optimization without normal label supervision, a mathematically closed imaging model must be constructed such that all predicted variables are mapped to the observed image through a unified forward rendering process.

To this end, we unify the rendering process as a mapping from the set of 3D Gaussian parameters $\{G_i\}$ to the image space. For each Gaussian, its appearance is no longer treated as an independent color variable but is jointly determined by its albedo and local illumination. Following the radiance contribution formula defined in Equation~\ref{eq315}, the radiance of the $i$-th Gaussian at pixel $\boldsymbol{x}$ is $C_i(\boldsymbol{x}) = \boldsymbol{k}_{D,i} \odot \max(0,\boldsymbol{n}_{fuse,i}^\top\boldsymbol{l}) \cdot K_i(\boldsymbol{p}(\boldsymbol{x}))$, where $\boldsymbol{k}_{D,i}$ is the albedo, $\boldsymbol{l}$ is the illumination direction, and $\boldsymbol{p}(\boldsymbol{x})$ is the 3D spatial point. This formulation establishes a direct differentiable link from 3D Gaussian parameters, through the fused normal and albedo, to per-pixel radiance.

On this basis, the pixel-wise rendering image is obtained through the 3DGS rasterization process. Let $T_i(\boldsymbol{x})$ denote the cumulative transmittance and $\alpha_i(\boldsymbol{x})$ denote the projected opacity of the $i$-th Gaussian at this pixel. The opacity $\alpha_i(\boldsymbol{x})$ is derived from $\sigma_i$ and $\Sigma_i$ following the standard Gaussian projection formulation. The visibility weight is defined as $w_i(\boldsymbol{x}) = T_i(\boldsymbol{x}) \alpha_i(\boldsymbol{x})$. The rendered image is defined as:
\begin{equation}
    \label{eq342}
    I_{\mathrm{pred}}(\boldsymbol{x})=\sum_{i=1}^N w_i(\boldsymbol{x}) C_i(\boldsymbol{x}) + \boldsymbol{I}_{\mathrm{ambient}},
\end{equation}
where $N$ is the total number of Gaussian primitives, $\boldsymbol{I}_{\mathrm{ambient}}$ denotes a constant ambient light term modeling global background illumination with a fixed low intensity, and the transmittance is given by:
\begin{equation}
    \label{eq343}
    T_i(\boldsymbol{x})=\prod_{j<i}(1-\alpha_j(\boldsymbol{x})).
\end{equation}

After obtaining the forward rendering result, we define the photometric consistency loss as:
\begin{equation}
    \label{eq344}
    \mathcal{L}_{\mathrm{photo}} = \|I_{\mathrm{pred}} - I\|_1.
\end{equation}

To further constrain the geometric structure, we introduce a scale regularization term:
\begin{equation}
    \label{eq345}
    \mathcal{L}_{\mathrm{scale}} = \sum_i \left(\lambda_{\mathrm{min}}^i + \gamma \lambda_{\mathrm{max}}^i\right),
\end{equation}
where $\lambda_{\mathrm{min}}^i$ and $\lambda_{\mathrm{max}}^i$ are eigenvalues of $\Sigma_i$ and $\gamma$ is a positive balance hyperparameter. This regularization prevents Gaussian primitives from degenerating into infinitely thin flats, where $\lambda_{\mathrm{min}} \to 0$ fails to define a well-posed surface normal, or excessively elongated ellipsoids, where $\lambda_{\mathrm{max}} \to \infty$ introduces numerical instability, thus maintaining a well-conditioned geometric representation throughout optimization.

We also introduce a normal consistency constraint that encourages the refined normal to stay close to the geometric normal, but only in regions where the gating weight is small, where the refinement is expected to be active. This adaptive weighting prevents the constraint from forcing the diffusion output to collapse to the geometric prior in areas where the gating already favors geometry. The loss is defined as:
\begin{equation}
    \label{eq346}
    \mathcal{L}_{\mathrm{normal}} = \left\| \boldsymbol{n}_{3dgs} - \boldsymbol{n}_{\mathrm{diff}} \right\|_1 \odot (1 - g),
\end{equation}
where $g$ is the spatially varying gating weight from Equation~\ref{eq333}, and $\odot$ denotes element-wise multiplication. This loss encourages consistency between the geometric and refined normal estimates: when the two branches diverge, the gradient drives $\boldsymbol{n}_{\mathrm{diff}}$ toward $\boldsymbol{n}_{3dgs}$, ensuring the refinement does not deviate arbitrarily from the geometric prior. The $(1-g)$ factor relaxes this constraint in texture-rich regions where the diffusion branch is expected to contribute detail corrections beyond the geometric estimate. Thus, $\mathcal{L}_{\mathrm{normal}}$ acts as a soft regularizer rather than a hard consistency constraint: it prevents $\boldsymbol{n}_{\mathrm{diff}}$ from deviating arbitrarily far from $\boldsymbol{n}_{3dgs}$ when the gating is uncertain ($g \to 0.5$), while allowing substantial corrections when the gating confidently favors the diffusion branch ($g \to 0$).

To enforce consistency between the learnable principal direction $\boldsymbol{d}$ and the geometric normal, we add a directional regularization that focuses on orientation rather than magnitude, explicitly normalizing both vectors to unit length:
\begin{equation}
    \label{eq347}
    \mathcal{L}_{\mathrm{dir}} = \sum_i \left\| \frac{\boldsymbol{d}_i}{\|\boldsymbol{d}_i\|_2} - \frac{\boldsymbol{n}_{3dgs,i}}{\|\boldsymbol{n}_{3dgs,i}\|_2} \right\|_2^2.
\end{equation}

To mitigate the inherent scale ambiguity between albedo and shading, we rely on two implicit regularizers: the Gaussian opacity $\sigma_i$ controls the contribution of each primitive in the splatting accumulation, and the multi-view consistency enforced by the reprojection loss across different viewing angles during training encourages a physically plausible decomposition. This avoids the need for explicit albedo priors while maintaining stable optimization.

The final loss is:
\begin{equation}
    \label{eq348}
    \mathcal{L}_{\mathrm{total}} = \mathcal{L}_{\mathrm{photo}} + \lambda_{\mathrm{scale}}\mathcal{L}_{\mathrm{scale}} + \lambda_{\mathrm{normal}}\mathcal{L}_{\mathrm{normal}} + \lambda_{\mathrm{dir}}\mathcal{L}_{\mathrm{dir}},
\end{equation}
where $\lambda_{\mathrm{scale}}$, $\lambda_{\mathrm{normal}}$ and $\lambda_{\mathrm{dir}}$ are balance hyperparameters.
This formulation establishes a joint optimization from image space to 3D geometry and back, ensuring stable convergence without ground-truth supervision.
More details about Implementation Details and Training Details are provided in Sections~\ref{secs1} in the Supplementary Material.

\section{Experiments and Analysis}\label{sec4}

\subsection{Datasets and Evaluation Metrics}\label{sec41}
To verify our method's effectiveness, we conduct weakly supervised experiments using only RGB-3D model paired data, with no pixel-level normal labels or dense point cloud supervision, adhering strictly to the low-label constraint.

\textbf{Training Data.}
We train on the Objaverse dataset~\cite{Deitke2023Objaverse}, which contains 80k diverse individual 3D object models, rendering RGB images from random viewpoints under a fixed directional light $\boldsymbol{l} = [0, 0, 1]^\top$ with $\|\boldsymbol{l}\|_2=1$ to form training pairs. Each training sample consists of an RGB image rendered from a randomly sampled viewpoint and its corresponding 3D mesh. The same lighting is used for training and testing to ensure consistent photometric constraints, and no normal ground-truth is accessed during optimization.

\textbf{Testing Data.}
Evaluation is performed on three object-level benchmarks: Google Scanned Objects~\cite{Downs2022GoogleScanned} with $1{,}030$ daily objects for general performance, Omniobject3D~\cite{Wu2023OmniObject3D} with 6k+ complex objects for fine-detail testing and Wonder3D~\cite{Long2024Wonder3D} with $100$ real-world images for qualitative cross-domain visualization only. Normal ground-truth is rendered with consistent camera and lighting settings, with zero overlap between training and test sets to avoid data leakage.

\textbf{Evaluation Metrics.}
We adopt standard normal estimation metrics: Mean Angle Error (MAE), Median Angle Error (MedAE) and three threshold-based accuracy metrics at $11.25^\circ$, $22.5^\circ$ and $30^\circ$, following widely adopted evaluation protocols in the literature. Lower MAE and MedAE values with higher accuracy scores indicate better performance. All quantitative results are reported as mean$\pm$std over three independent runs for statistical reliability.

\textbf{Implementation Settings.}
All experiments follow unified hardware and software setups, with an NVIDIA RTX 4090 GPU, PyTorch 2.1.0 and CUDA 11.8. We use the two-stage training strategy detailed in the training details section, with fixed hyperparameters across all runs. All baselines are retrained on our Objaverse split to achieve their optimal performance, following each method's original training protocol. Baselines are split into dedicated single-image normal estimation methods and generative 3D reconstruction methods for clear comparison.
More details about Implementation Details and Training Details see Section~\ref{secs1} in the Supplementary Material.

\subsection{Comparative Experiments}\label{sec42}

To comprehensively validate the effectiveness of our method, we conduct both quantitative and qualitative comparisons with state-of-the-art methods across all test datasets, adhering to a unified experimental protocol.

We compare with thirteen recent representative methods spanning diverse technical paradigms: generative normal estimation methods (StableNormal~\cite{Ye2024StableNormal}, GenPercept~\cite{Xu2025GenPercept}), diffusion-based normal predictors (Marigold-Normals~\cite{Ke2024Marigold}, GeoWizard~\cite{Ye2025GeoSplatting}), discriminative deterministic geometry methods (Metric3D v2~\cite{Hu2024Metric3Dv2}, DSINE~\cite{Bae2024DSINE}), multi-view reconstruction methods (One-2-3-45++~\cite{Liu2024One2345++}), neural light transport methods (Neural LightRig~\cite{He2025NeuralLightRig}), geometry-focused generative methods (Wonder3D++~\cite{Yang2026Wonder3D}, FE2E~\cite{Wang2025FE2E}, Hyden~\cite{Zhang2026Hyden}, URGT~\cite{Cui2026URGT}) and inverse rendering-based methods (RoSE~\cite{Li2026RoSE}). All methods are evaluated under their standard inference protocols using official implementations where available.

\begin{table*}[!htbp]
\centering
\caption{Quantitative Comparison with State-of-the-Art Methods.}
\label{tab:comparison}
\setlength{\tabcolsep}{2pt}
    \begin{tabular}{lllccccc}
    \toprule
    \multirow{2}{*}{Method} & \multirow{2}{*}{Venue} & \multirow{2}{*}{Supervisors} & MAE$\downarrow$ ($^\circ$) & MedAE$\downarrow$ ($^\circ$) & Acc@11.25$^\circ$$\uparrow$  (\%) & Acc@22.5$^\circ$$\uparrow$  (\%) & Acc@30$^\circ$$\uparrow$  (\%) \\
    & & & (mean$\pm$std) & (mean$\pm$std) & (mean$\pm$std) & (mean$\pm$std) & (mean$\pm$std) \\
    \midrule
    Metric3D v2~\cite{Hu2024Metric3Dv2} & TPAMI 2024 & Full & 14.1$\pm$0.3 & 11.5$\pm$0.4 & 61.0$\pm$0.2 & 77.2$\pm$0.5 & 83.1$\pm$0.1 \\
    One-2-3-45++~\cite{Liu2024One2345++} & CVPR 2024 & Full & 15.7$\pm$0.2 & 12.9$\pm$0.5 & 55.4$\pm$0.4 & 73.8$\pm$0.3 & 79.6$\pm$0.2 \\
    DSINE~\cite{Bae2024DSINE} & CVPR 2024 & Full & 14.8$\pm$0.4 & 12.1$\pm$0.2 & 58.9$\pm$0.3 & 75.6$\pm$0.5 & 81.4$\pm$0.4 \\
    Marigold-Normals~\cite{Ke2024Marigold} & CVPR 2024 & Weakly & 16.8$\pm$0.1 & 13.9$\pm$0.3 & 54.1$\pm$0.5 & 71.2$\pm$0.2 & 77.6$\pm$0.4 \\
    GeoWizard~\cite{Ye2025GeoSplatting} & ECCV 2024 & Weakly & 15.9$\pm$0.5 & 13.1$\pm$0.1 & 55.8$\pm$0.3 & 73.0$\pm$0.4 & 79.1$\pm$0.2 \\
    StableNormal~\cite{Ye2024StableNormal} & SIGGRAPH 2024 & Weakly & 15.3$\pm$0.2 & 12.4$\pm$0.4 & 57.3$\pm$0.5 & 74.5$\pm$0.1 & 80.2$\pm$0.3 \\
    URGT~\cite{Cui2026URGT} & arXiv 2025 & Full & 18.2$\pm$0.4 & 15.6$\pm$0.3 & 48.9$\pm$0.2 & 68.2$\pm$0.5 & 73.5$\pm$0.1 \\
    GenPercept~\cite{Xu2025GenPercept} & ICLR 2025 & Weakly & 14.6$\pm$0.3 & 11.8$\pm$0.5 & 60.1$\pm$0.1 & 76.8$\pm$0.4 & 82.5$\pm$0.2 \\
    Neural LightRig~\cite{He2025NeuralLightRig} & CVPR 2025 & Self & 16.8$\pm$0.2 & 13.9$\pm$0.4 & 54.0$\pm$0.3 & 71.0$\pm$0.5 & 77.5$\pm$0.4 \\
    Wonder3D++~\cite{Yang2026Wonder3D} & TPAMI 2026 & Weakly & 16.4$\pm$0.5 & 13.8$\pm$0.2 & 53.1$\pm$0.4 & 71.4$\pm$0.3 & 77.8$\pm$0.1 \\
    FE2E~\cite{Wang2025FE2E} & CVPR 2026 & Weakly & 14.3$\pm$0.1 & 11.7$\pm$0.3 & 59.5$\pm$0.5 & 76.8$\pm$0.2 & 82.7$\pm$0.4 \\
    RoSE~\cite{Li2026RoSE} & ICLR 2026 & Self & 14.4$\pm$0.4 & 12.1$\pm$0.2 & 58.4$\pm$0.3 & 75.8$\pm$0.5 & 81.2$\pm$0.2 \\
    Hyden~\cite{Zhang2026Hyden} & ICLR 2026 & Weakly & 15.1$\pm$0.2 & 12.6$\pm$0.5 & 56.3$\pm$0.1 & 73.9$\pm$0.4 & 80.1$\pm$0.3 \\
    \midrule
    \textbf{CLONE (Ours)} & --- & Weakly & \textbf{13.2$\pm$0.2} & \textbf{10.7$\pm$0.2} & \textbf{63.2$\pm$0.3} & \textbf{79.7$\pm$0.3} & \textbf{84.6$\pm$0.2} \\
    \bottomrule
    \end{tabular}
    \vspace{-16pt}
\end{table*}

As shown in Table~\ref{tab:comparison}, under the weakly-supervised paradigm without normal labels, CLONE achieves the best performance across all metrics. CLONE outperforms the strongest fully-supervised baseline Metric3D v2 and the strongest weakly-supervised baseline FE2E in mean angle error. All baselines were retrained on our Objaverse split following their original protocols: fully-supervised methods such as Metric3D v2, DSINE, One-2-3-45++ and URGT used rendered ground-truth normals, while weakly-supervised and self-supervised methods used only image-3D pairs without normal labels, matching their original training paradigms. CLONE was trained on the same image-3D pairs without using the normal ground truth. The consistent advantage across all accuracy thresholds confirms CLONE's superior geometric reasoning capability under weak supervision.

\begin{figure}[!htbp]
    \centering
    \includegraphics[width=1.0\linewidth]{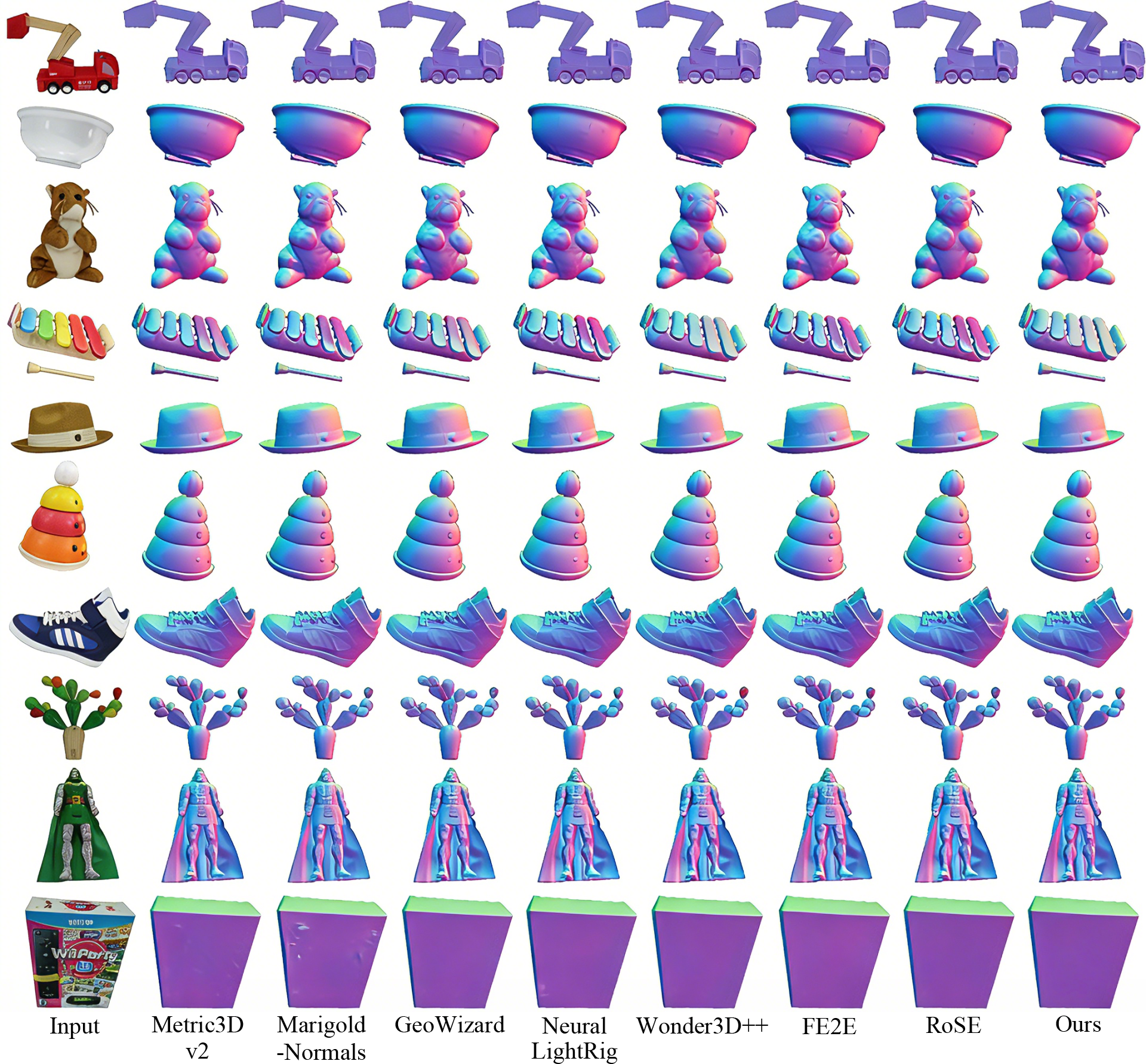}
    \caption{Qualitative Comparison with SOTAs on GSO Dataset.}
    \label{fig-sota1}
\end{figure}

\begin{figure}[!htbp]
    \centering
    \includegraphics[width=1.0\linewidth]{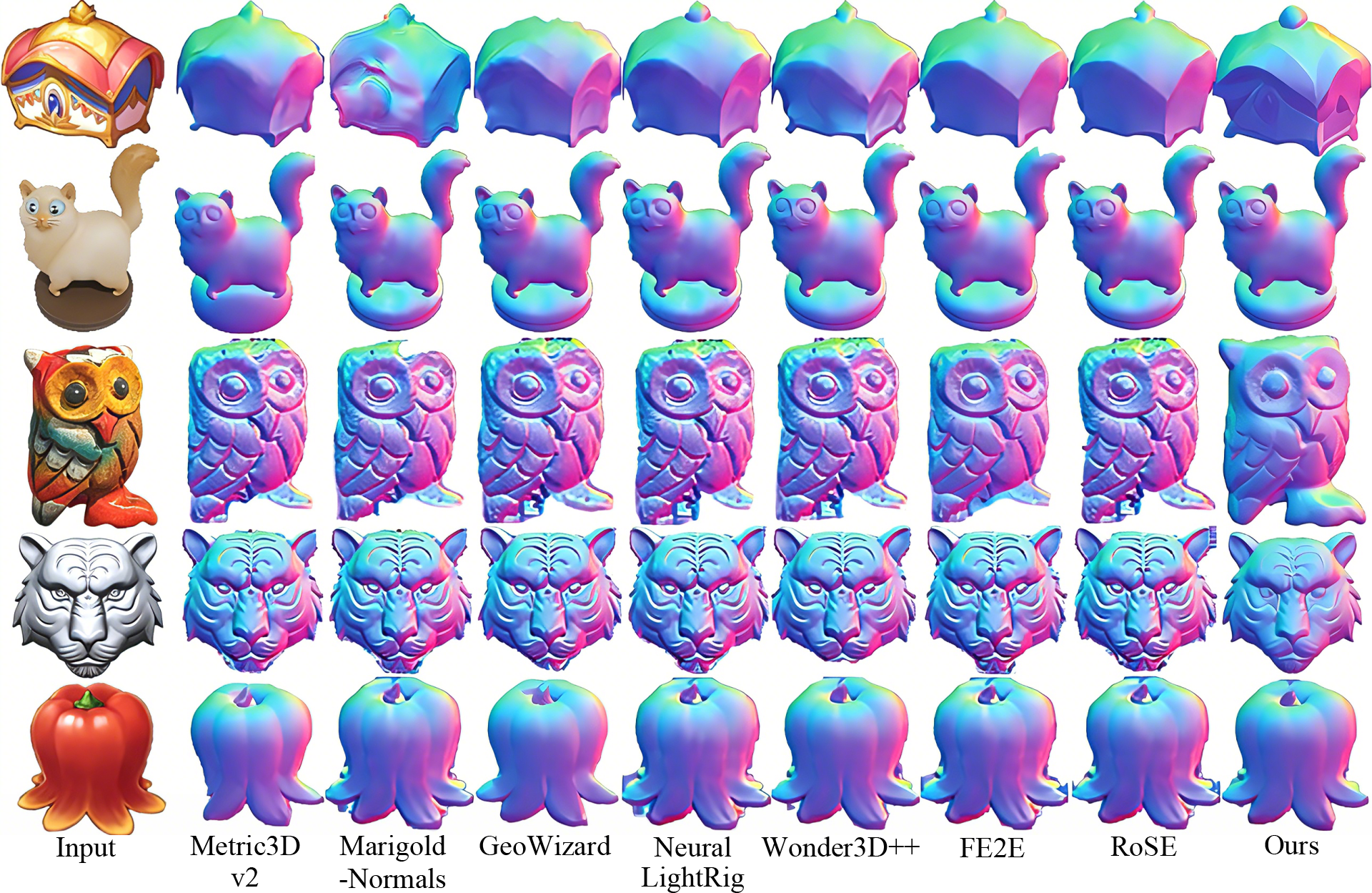}
    \caption{Qualitative Comparison with SOTAs on Omniobject3D Dataset.}
    \label{fig-sota2}
\end{figure}

\begin{figure}[!htbp]
    \centering
    \includegraphics[width=1.0\linewidth]{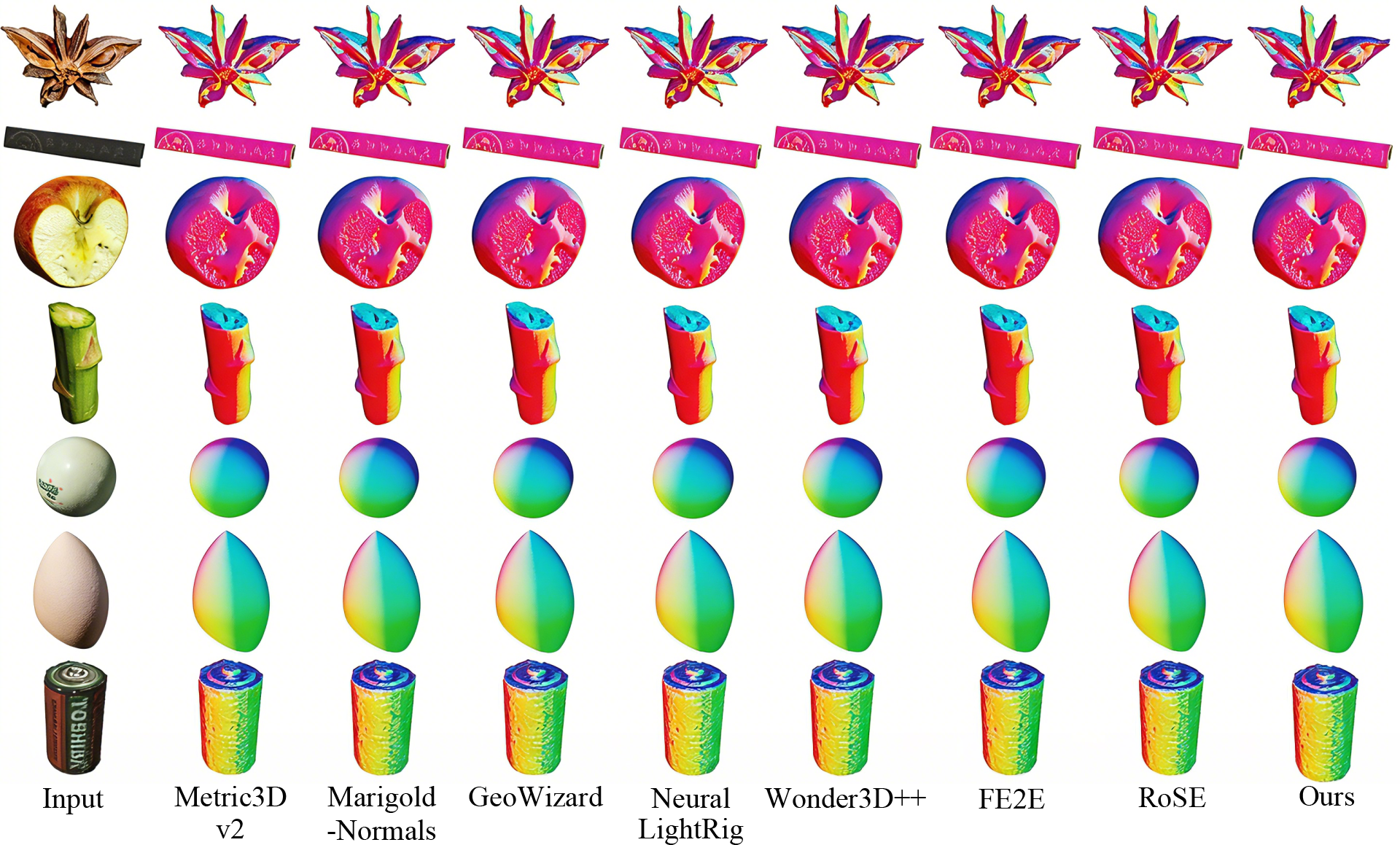}
    \caption{Qualitative Comparison with SOTAs on Wonder3D Dataset.}
    \label{fig-sota3}
\end{figure}

Qualitative comparisons across diverse object categories are presented in Figures~\ref{fig-sota1}--\ref{fig-sota3}, covering smooth primitive shapes, textured objects, thin and elongated structures and complex topologies.

For smooth, low-texture objects such as bowls, spheres, and eggs (Fig.~\ref{fig-sota1}, \ref{fig-sota3}), generative methods over-smooth, erasing subtle cues and producing flat normal maps, whereas our method preserves smooth transitions, sharp edges, and accurate curvature. On high-frequency textured objects like keyboards, owls, tiger heads, and leaves (Fig.~\ref{fig-sota1}, \ref{fig-sota2}), competing methods introduce artifacts or erase key contours and patterns; our method recovers fine geometric details without artifacts, consistent with our highest quantitative accuracy. For thin, elongated structures such as shoes, cactus branches, and bamboo (Fig.~\ref{fig-sota1}, \ref{fig-sota3}), competing methods often break normal consistency across narrow surfaces, while our method maintains continuous normals and full silhouettes even in high-curvature regions. On complex topologies with sharp junctions like excavators and house models (Fig.~\ref{fig-sota1}, \ref{fig-sota2}), competing methods struggle with consistency across connected parts. Our method clearly resolves part boundaries and junctions with consistent orientation.

\subsection{Ablation Study}\label{sec43}
To rigorously validate the necessity and synergistic effectiveness of each core component in our framework, we conduct a comprehensive controlled ablation study. All ablation variants strictly adopt identical training settings, random seeds and hyperparameters to ensure a fair comparison.

\begin{table}[!htbp]
\centering
\caption{Ablation Results of Core Modules. Metrics are Mean$\pm$std over $3$ Runs.}
\label{tab:ablation_modules}
\setlength{\tabcolsep}{2pt}
    \begin{tabular}{lccccc}
    \toprule
    \makecell{Variant} & \makecell{MAE$\downarrow$} & \makecell{MedAE$\downarrow$} & \makecell{Acc@11.25$^\circ$\\$\uparrow$} & \makecell{Acc@22.5$^\circ$\\$\uparrow$} & \makecell{Acc@30$^\circ$\\$\uparrow$} \\
    \midrule
    \multirow{2}{*}{\makecell[l]{w/o 3DGS \\ Initialization}} & 15.7$\pm$0.5 & 12.9$\pm$0.4 & 56.3$\pm$0.5 & 74.1$\pm$0.5 & 79.5$\pm$0.5 \\
    & & & & & \\
    \multirow{2}{*}{\makecell[l]{w/o Modulation \\ Kernel}} & 13.9$\pm$0.3 & 11.5$\pm$0.3 & 60.6$\pm$0.4 & 77.4$\pm$0.4 & 82.7$\pm$0.3 \\
    & & & & & \\
    \multirow{2}{*}{\makecell[l]{w/o Cross-Domain \\ Alignment}} & 13.9$\pm$0.3 & 11.4$\pm$0.3 & 60.8$\pm$0.4 & 77.6$\pm$0.4 & 82.9$\pm$0.3 \\
    & & & & & \\
    \multirow{2}{*}{\makecell[l]{w/o Gating \\ Fusion}} & 13.6$\pm$0.3 & 11.1$\pm$0.2 & 61.7$\pm$0.4 & 78.2$\pm$0.3 & 83.5$\pm$0.3 \\
    & & & & & \\
    \multirow{2}{*}{\makecell[l]{w/o Diffusion \\ Refinement}} & 13.4$\pm$0.2 & 10.9$\pm$0.2 & 62.5$\pm$0.3 & 78.9$\pm$0.3 & 84.0$\pm$0.3 \\
    & & & & & \\
    \textbf{Full Model} & \textbf{13.2$\pm$0.2} & \textbf{10.7$\pm$0.2} & \textbf{63.2$\pm$0.3} & \textbf{79.7$\pm$0.3} & \textbf{84.6$\pm$0.2} \\
    \bottomrule
    \end{tabular}
\end{table}

\begin{figure}
    \centering
    \includegraphics[width=0.8\linewidth]{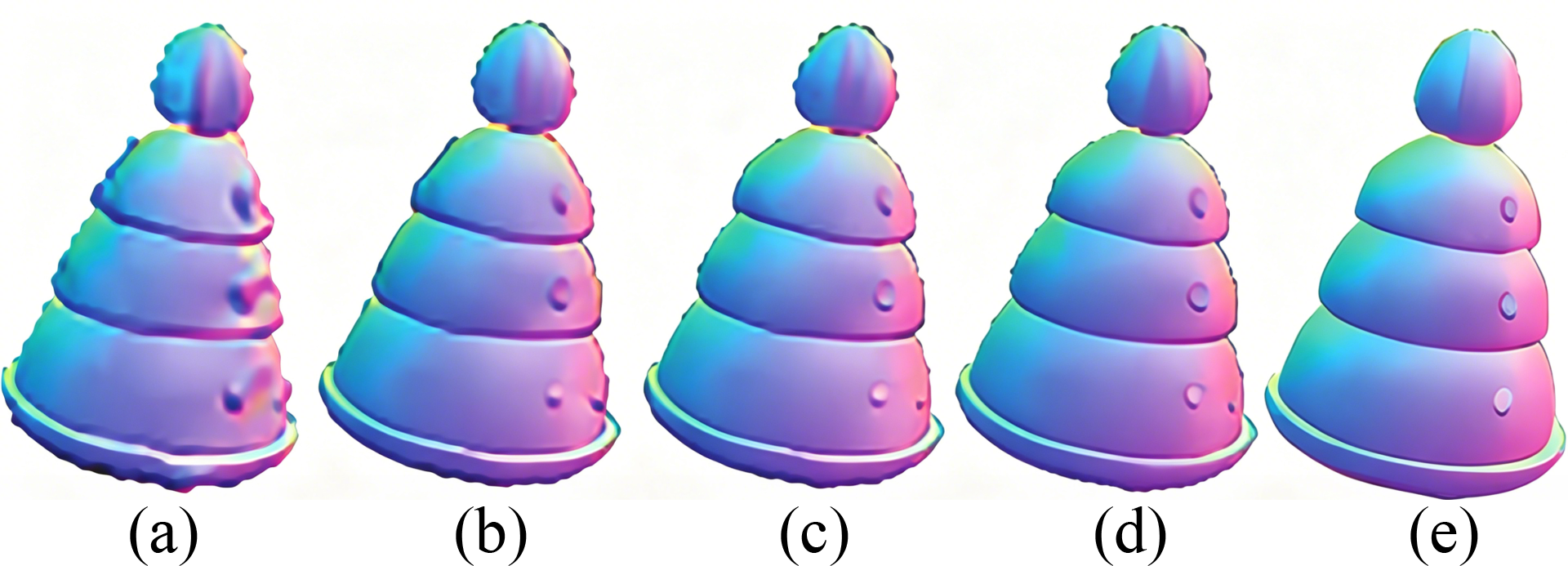}
    \caption{Comparative Results about Our Proposed Components. (a) w/o 3DGS Initialization. (b) w/o Cross-Domain Alignment. (c) w/o Gating Fusion. (d) w/o Diffusion Refinement. (e) Full Model.}
    \label{fig-ablation}
\end{figure}

\textbf{Core Components.}
As presented in Table~\ref{tab:ablation_modules} and Figure~\ref{fig-ablation}, removing any core module leads to significant performance degradation across all metrics and confirms their indispensable functions. Eliminating the 3DGS geometric initialization module results in the sharpest performance decline, with a substantial increase in MAE and a marked drop in accuracy, since it provides explicit 3D geometric priors to alleviate the ill-posed problem of monocular normal estimation and acts as the fundamental geometric backbone of the whole framework. Removing the learnable modulation kernel elevates the MAE, which verifies that the oscillatory kernel can capture valid high-frequency geometric information beyond simple diffuse reflectance constraints and function as a structural enhancement term for micro-geometry variation modeling. Meanwhile, discarding the cross-domain alignment module also increases the MAE, revealing that these two modules deliver comparable performance gains and compensate for distinct geometric errors, with the former excavating intrinsic micro-geometry and the latter facilitating effective multi-modal representation fusion, so the lack of either will trigger similar severe performance deterioration. The adaptive gating fusion module assigns dynamic feature weights based on local surface complexity, and its removal induces feature redundancy and noise disturbance with an increased MAE, a performance level consistent with that of static fusion schemes in Section~\ref{secs21} of the Supplementary Material, proving that adaptive weighting is essential to escape inferior non-adaptive fusion performance. In addition, removing the diffusion refinement module restricts the model to unoptimized raw 3DGS normals and weakens fine detail restoration ability, further validating its vital effect on precise normal refinement.

\textbf{Loss Weight Sensitivity and Loss Terms.}
We perform a grid search over the key loss balancing hyperparameters $\lambda_{\mathrm{scale}}$ and $\lambda_{\mathrm{normal}}$ to validate the robustness of our chosen settings. Table~\ref{tab:hyperparam} shows that the optimal configuration yields the best MAE, while nearby values exhibit only mild degradation, confirming that the training is not highly sensitive to these choices.
Table~\ref{tab:ablation_losses} quantifies the contribution of each loss component. Using only photometric loss $\mathcal{L}_{\mathrm{photo}}$ (trained in a single-stage joint optimization without $\mathcal{L}_{\mathrm{scale}}$ in any phase) leads to catastrophic performance collapse, as pure photometric supervision cannot constrain 3D geometric consistency, causing Gaussian primitive collapse and severe normal distortion. Removing the scale regularization $\mathcal{L}_{\mathrm{scale}}$ results in degenerate Gaussian parameters and unstable optimization, leading to a higher MAE. Removing the normal consistency constraint $\mathcal{L}_{\mathrm{normal}}$ breaks the alignment between diffusion-refined normals and 3DGS geometric normals, introducing artifacts and raising the MAE. The full loss combination achieves optimal performance via complementary geometric and photometric constraints, validating the synergistic design of the loss function.
More Experiments are provided in Section~\ref{secs2} in the Supplementary Material.

\begin{table}[!htbp]
\centering
\caption{Hyperparameter Sensitivity for Loss Weights $\lambda_{\mathrm{scale}}$ and $\lambda_{\mathrm{normal}}$, MAE$\downarrow$ ($^\circ$). Metrics are Mean$\pm$std over $3$ Runs.}
\label{tab:hyperparam}
\scriptsize
\setlength{\tabcolsep}{6pt}
    \begin{tabular}{lcccc}
    \toprule
    $\lambda_{\mathrm{scale}}$ & \multicolumn{4}{c}{$\lambda_{\mathrm{normal}}$} \\
    \cmidrule(lr){2-5}
    & 0.1 & 0.3 & \textbf{0.5} & 0.7 \\
    \midrule
    0.05  & 14.1$\pm$0.3 & 13.8$\pm$0.4 & 13.5$\pm$0.3 & 13.7$\pm$0.2 \\
    \textbf{0.1}  & 13.7$\pm$0.3 & 13.4$\pm$0.2 & \textbf{13.2$\pm$0.2} & 13.5$\pm$0.3 \\
    0.2  & 13.9$\pm$0.2 & 13.6$\pm$0.3 & 13.4$\pm$0.3 & 13.6$\pm$0.4 \\
    \bottomrule
    \end{tabular}
\end{table}

\begin{table}[!htbp]
\centering
\caption{Ablation Results of Loss Functions. Metrics are Mean$\pm$std over $3$ Runs.}
\label{tab:ablation_losses}
\scriptsize
\setlength{\tabcolsep}{1pt}
    \begin{tabular}{lccccc}
    \toprule
    \multirow{2}{*}{Variant}
    & MAE$\downarrow$
    & MedAE$\downarrow$
    & Acc@11.25$^\circ$$\uparrow$
    & Acc@22.5$^\circ$$\uparrow$
    & Acc@30$^\circ$$\uparrow$ \\
    & (mean$\pm$std)
    & (mean$\pm$std)
    & (mean$\pm$std)
    & (mean$\pm$std)
    & (mean$\pm$std) \\
    \midrule
    $\mathcal{L}_{\mathrm{photo}}$ only & 35.7$\pm$0.5 & 28.4$\pm$0.5 & 12.3$\pm$0.5 & 25.6$\pm$0.5 & 31.2$\pm$0.5 \\
    w/o $\mathcal{L}_{\mathrm{scale}}$ & 15.5$\pm$0.5 & 12.6$\pm$0.4 & 58.9$\pm$0.5 & 73.5$\pm$0.4 & 80.2$\pm$0.4 \\
    w/o $\mathcal{L}_{\mathrm{normal}}$ & 13.9$\pm$0.3 & 11.2$\pm$0.2 & 60.9$\pm$0.4 & 76.8$\pm$0.3 & 81.9$\pm$0.3 \\
    Full loss & \textbf{13.2$\pm$0.2} & \textbf{10.7$\pm$0.2} & \textbf{63.2$\pm$0.3} & \textbf{79.7$\pm$0.3} & \textbf{84.6$\pm$0.2} \\
    \bottomrule
    \end{tabular}
\end{table}

\subsection{Robustness Tests}\label{sec44}
We conduct comprehensive robustness evaluations under challenging inference conditions, including input noise, illumination variations and viewpoint changes. All tests follow the same settings as the main experiments to ensure consistency.

\textbf{Input Noise Robustness.}
We test model stability under additive Gaussian noise with standard deviations $\sigma_{\mathrm{noise}} \in \{0,5,10,15\}$. As shown in Table~\ref{tab:robustness_noise}, our method maintains exceptional stability: MAE increases by only $6.0^\circ$ from $\sigma_{\mathrm{noise}}=0$ to $\sigma_{\mathrm{noise}}=15$, whereas the most affected baseline, Marigold-Normals, suffers a $13.3^\circ$ MAE surge over the same noise range. The explicit 3DGS geometric representation and differentiable rasterization gradient aggregation effectively suppress noise interference, outperforming pure 2D regression methods in noisy environments.

\begin{table}[!htbp]
\centering
\caption{Robustness to Additive Gaussian Noise, MAE$\downarrow$ ($^\circ$). Metrics are Mean$\pm$std over $3$ Runs.}
\label{tab:robustness_noise}
\scriptsize
\setlength{\tabcolsep}{2pt}
    \begin{tabular}{lcccc}
    \toprule
    Method & $\sigma_{\mathrm{noise}}=0$ & $\sigma_{\mathrm{noise}}=5$ & $\sigma_{\mathrm{noise}}=10$ & $\sigma_{\mathrm{noise}}=15$ \\
    \midrule
    Metric3D v2~\cite{Hu2024Metric3Dv2}      & 14.1$\pm$0.3 & 15.8$\pm$0.3 & 18.2$\pm$0.4 & 22.3$\pm$0.5 \\
    One-2-3-45++~\cite{Liu2024One2345++}    & 15.7$\pm$0.4 & 18.5$\pm$0.5 & 22.7$\pm$0.5 & 28.4$\pm$0.5 \\
    DSINE~\cite{Bae2024DSINE}                & 14.8$\pm$0.3 & 16.6$\pm$0.4 & 19.1$\pm$0.5 & 23.5$\pm$0.5 \\
    Marigold-Normals~\cite{Ke2024Marigold}   & 16.8$\pm$0.4 & 19.9$\pm$0.5 & 24.3$\pm$0.5 & 30.1$\pm$0.5 \\
    GeoWizard~\cite{Ye2025GeoSplatting}         & 15.9$\pm$0.3 & 18.9$\pm$0.4 & 23.1$\pm$0.5 & 28.9$\pm$0.5 \\
    StableNormal~\cite{Ye2024StableNormal}   & 15.3$\pm$0.4 & 18.0$\pm$0.5 & 22.0$\pm$0.5 & 27.2$\pm$0.5 \\
    URGT~\cite{Cui2026URGT}                  & 18.2$\pm$0.4 & 20.7$\pm$0.4 & 24.5$\pm$0.5 & 29.8$\pm$0.5 \\
    GenPercept~\cite{Xu2025GenPercept}    & 14.6$\pm$0.3 & 16.9$\pm$0.4 & 20.5$\pm$0.5 & 25.6$\pm$0.5 \\
    Neural LightRig~\cite{He2025NeuralLightRig} & 16.8$\pm$0.4 & 19.2$\pm$0.4 & 23.0$\pm$0.5 & 28.1$\pm$0.5 \\
    Wonder3D++~\cite{Yang2026Wonder3D}       & 16.4$\pm$0.4 & 19.5$\pm$0.5 & 23.9$\pm$0.5 & 29.5$\pm$0.5 \\
    FE2E~\cite{Wang2025FE2E}                & 14.3$\pm$0.3 & 16.1$\pm$0.3 & 18.9$\pm$0.4 & 23.8$\pm$0.5 \\
    RoSE~\cite{Li2026RoSE}                   & 14.4$\pm$0.3 & 16.3$\pm$0.3 & 19.2$\pm$0.4 & 24.1$\pm$0.5 \\
    Hyden~\cite{Zhang2026Hyden}                 & 15.1$\pm$0.3 & 17.5$\pm$0.4 & 21.0$\pm$0.5 & 26.3$\pm$0.5 \\
    \textbf{CLONE (Ours)}                    & \textbf{13.2$\pm$0.2} & \textbf{14.8$\pm$0.2} & \textbf{16.5$\pm$0.3} & \textbf{19.2$\pm$0.4} \\
    \bottomrule
    \end{tabular}
\end{table}

\begin{figure}
    \centering
    \includegraphics[width=1.0\linewidth]{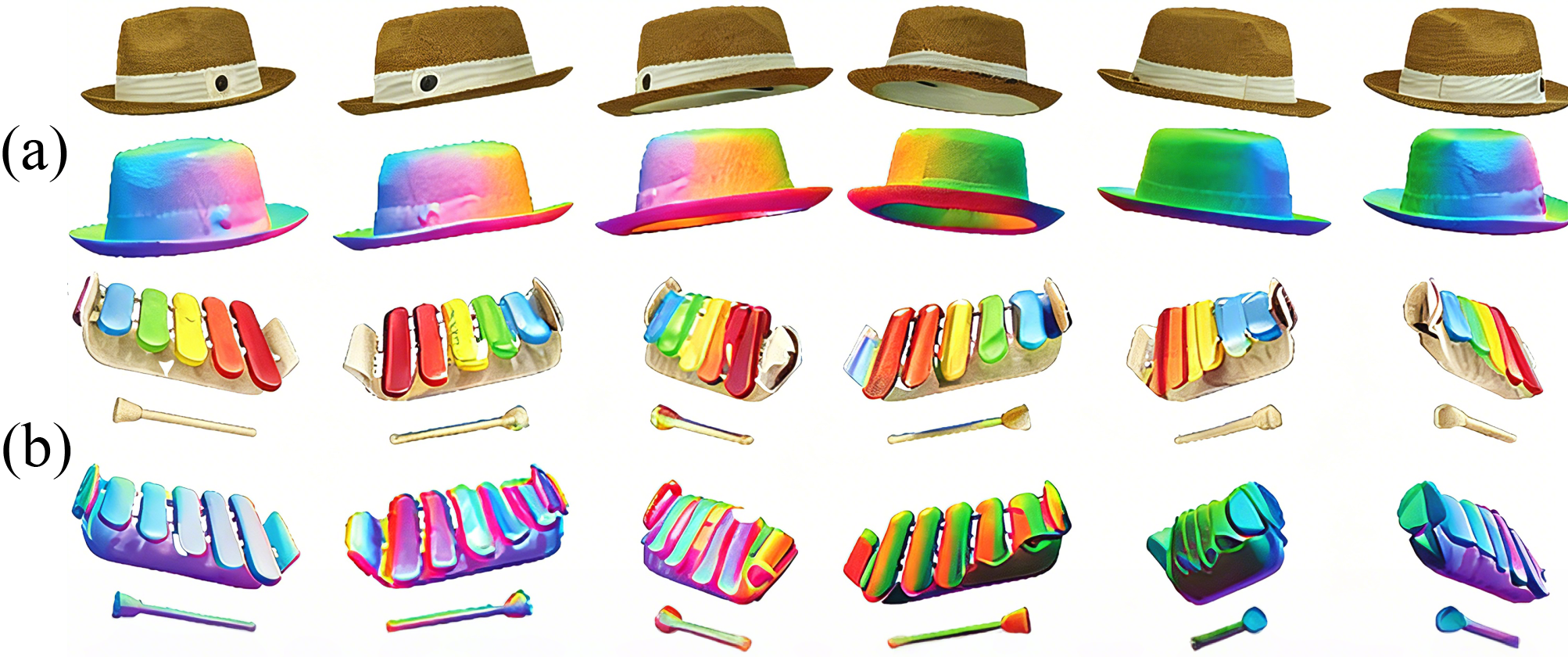}
    \caption{Robustness Case: Multi-Viewpoint.}
    \label{fig-robustcase1}
\end{figure}

\textbf{Lighting \& Viewpoint Robustness.}
Tables~\ref{tab:robustness_light},~\ref{tab:robustness_viewpoint} and~\ref{fig-robustcase1} present comprehensive performance comparisons under three illumination directions (front, side, top) and different representative viewpoints (front at $0^\circ$, side at $45^\circ$, oblique at $70^\circ$), including all competing methods evaluated in the main comparison. Our method consistently achieves the lowest MAE across all settings, exhibiting superior generalization to environmental variations. Generative methods exhibit a measurable but moderate decline under viewpoint shifts, while discriminative methods also show non-negligible degradation under non-training viewpoints, as both paradigms rely on appearance cues that vary with viewing direction. In contrast, our joint geometric optimization framework effectively decouples normal estimation from appearance variations, maintaining stable and accurate predictions across lighting and viewpoint variations at the single-object level.

\begin{table}[!htbp]
\centering
\caption{Robustness to Lighting Variations (MAE$\downarrow$,$^\circ$, Mean$\pm$std over $3$ Runs)}
\label{tab:robustness_light}
\scriptsize
\setlength{\tabcolsep}{8pt}
    \begin{tabular}{lccc}
    \toprule
    Method & Front light & Side light & Top light \\
    \midrule
    Metric3D v2~\cite{Hu2024Metric3Dv2}      & 14.1$\pm$0.3 & 15.7$\pm$0.4 & 15.2$\pm$0.3 \\
    One-2-3-45++~\cite{Liu2024One2345++}    & 15.7$\pm$0.4 & 18.2$\pm$0.5 & 17.5$\pm$0.4 \\
    DSINE~\cite{Bae2024DSINE}                & 14.8$\pm$0.3 & 16.3$\pm$0.4 & 15.9$\pm$0.4 \\
    Marigold-Normals~\cite{Ke2024Marigold}   & 16.8$\pm$0.4 & 19.3$\pm$0.5 & 18.6$\pm$0.4 \\
    GeoWizard~\cite{Ye2025GeoSplatting}         & 15.9$\pm$0.3 & 18.3$\pm$0.4 & 17.6$\pm$0.4 \\
    StableNormal~\cite{Ye2024StableNormal}   & 15.3$\pm$0.4 & 17.5$\pm$0.5 & 16.8$\pm$0.4 \\
    URGT~\cite{Cui2026URGT}                  & 18.2$\pm$0.4 & 19.8$\pm$0.4 & 19.4$\pm$0.3 \\
    GenPercept~\cite{Xu2025GenPercept}    & 14.6$\pm$0.3 & 16.8$\pm$0.4 & 16.1$\pm$0.4 \\
    Neural LightRig~\cite{He2025NeuralLightRig} & 16.8$\pm$0.4 & 18.5$\pm$0.4 & 18.0$\pm$0.3 \\
    Wonder3D++~\cite{Yang2026Wonder3D}       & 16.4$\pm$0.4 & 18.9$\pm$0.5 & 18.2$\pm$0.4 \\
    FE2E~\cite{Wang2025FE2E}                & 14.3$\pm$0.3 & 15.9$\pm$0.3 & 15.4$\pm$0.3 \\
    RoSE~\cite{Li2026RoSE}                   & 14.4$\pm$0.3 & 16.1$\pm$0.3 & 15.6$\pm$0.3 \\
    Hyden~\cite{Zhang2026Hyden}                 & 15.1$\pm$0.3 & 16.9$\pm$0.4 & 16.4$\pm$0.3 \\
    \textbf{CLONE (Ours)}                    & \textbf{13.2$\pm$0.2} & \textbf{14.1$\pm$0.2} & \textbf{14.0$\pm$0.2} \\
    \bottomrule
    \end{tabular}
\end{table}

\begin{table}[!htbp]
\centering
\caption{Robustness to Viewpoint Changes, MAE$\downarrow$ ($^\circ$). Metrics are Mean$\pm$std over $3$ Runs.}
\label{tab:robustness_viewpoint}
\scriptsize
\setlength{\tabcolsep}{4pt}
    \begin{tabular}{lccc}
    \toprule
    Method & Front view ($0^\circ$) & Side view ($45^\circ$) & Oblique view ($70^\circ$) \\
    \midrule
    Metric3D v2~\cite{Hu2024Metric3Dv2}      & 14.1$\pm$0.3 & 15.3$\pm$0.3 & 16.5$\pm$0.4 \\
    One-2-3-45++~\cite{Liu2024One2345++}    & 15.7$\pm$0.4 & 18.0$\pm$0.4 & 19.5$\pm$0.5 \\
    DSINE~\cite{Bae2024DSINE}                & 14.8$\pm$0.3 & 16.0$\pm$0.4 & 17.3$\pm$0.4 \\
    Marigold-Normals~\cite{Ke2024Marigold}   & 16.8$\pm$0.4 & 19.3$\pm$0.4 & 21.0$\pm$0.5 \\
    GeoWizard~\cite{Ye2025GeoSplatting}         & 15.9$\pm$0.3 & 18.3$\pm$0.4 & 19.9$\pm$0.5 \\
    StableNormal~\cite{Ye2024StableNormal}   & 15.3$\pm$0.4 & 17.2$\pm$0.5 & 18.8$\pm$0.5 \\
    URGT~\cite{Cui2026URGT}                  & 18.2$\pm$0.4 & 19.5$\pm$0.4 & 20.9$\pm$0.4 \\
    GenPercept~\cite{Xu2025GenPercept}    & 14.6$\pm$0.3 & 16.5$\pm$0.4 & 17.9$\pm$0.4 \\
    Neural LightRig~\cite{He2025NeuralLightRig} & 16.8$\pm$0.4 & 18.5$\pm$0.3 & 19.8$\pm$0.4 \\
    Wonder3D++~\cite{Yang2026Wonder3D}       & 16.4$\pm$0.4 & 18.9$\pm$0.4 & 20.6$\pm$0.5 \\
    FE2E~\cite{Wang2025FE2E}                & 14.3$\pm$0.3 & 15.8$\pm$0.3 & 17.1$\pm$0.4 \\
    RoSE~\cite{Li2026RoSE}                   & 14.4$\pm$0.3 & 16.0$\pm$0.3 & 17.3$\pm$0.4 \\
    Hyden~\cite{Zhang2026Hyden}                 & 15.1$\pm$0.3 & 16.8$\pm$0.3 & 18.1$\pm$0.4 \\
    \textbf{CLONE (Ours)}                    & \textbf{13.2$\pm$0.2} & \textbf{13.8$\pm$0.2} & \textbf{14.5$\pm$0.2} \\
    \bottomrule
    \end{tabular}
\end{table}

\subsection{Efficiency Analysis}\label{sec45}

This section analyzes the efficiency of the proposed method through parameter distribution and practical inference performance.
The model scale is moderate among mainstream single-object normal estimation methods. Compared with lighter backbones, the selected backbone incurs a moderate parameter increase but yields noticeable accuracy improvement. The auxiliary cross-domain fusion and gating modules add only a small number of parameters while ensuring stable gradient propagation and complementary feature learning, which are vital for training stability and favorable predictions. Benefiting from one-step diffusion refinement, the method avoids heavy multi-step denoising costs and has strong acceleration potential. Its inference latency is higher than lightweight feed-forward and efficient generative models, mainly due to cross-domain feature interaction and spatial adaptive regulation required for high-precision optimization. In conclusion, the method achieves prominent performance gains at acceptable computational cost, and its balance between efficiency and accuracy suits practical high-precision visual normal estimation tasks.
More details are provided in Section~\ref{secs2} in the Supplementary Material.

\subsection{Discussions}\label{sec46}
While our method achieves strong performance, its limitations primarily arise from texture-geometry confusion, occlusion and slender structures, as well as non-Lambertian materials. Specifically, high texture gradients can mislead the diffusion branch, and specular reflections give rise to systematic errors. Furthermore, severe self-occlusion remains intractable due to the absence of explicit occlusion modeling. Even with the learnable modulation kernel incorporated, the simplified light interaction model still suffers from systematic errors on non-Lambertian surfaces, including glossy, metallic and transparent ones. High-texture surfaces can disrupt the model’s gating mechanism: when texture gradients dominate the input, the associated features will misguide the diffusion branch to mistake texture edges for geometric contours.
This work focuses solely on the task of single-object normal estimation and does not cover scene-level benchmarks, for scene-level challenges such as multi-object occlusion, complex environmental lighting and background separation fall outside the scope of the current research. Nevertheless, to address the aforementioned limitations, subsequent work will introduce a BRDF model with learnable roughness parameters to replace the existing Lambertian model while maintaining differentiability. Meanwhile, a self-supervised pre-training pipeline tailored for egocentric object-centric video data will be developed, which reduces the model's dependence on synthetic 3D model data by exploiting multi-view photometric consistency.
More Details are provided in Section~\ref{secs2} in the Supplementary Material.

\section{Conclusion}\label{sec5}

We proposed CLONE, a \textbf{C}ontinuous \textbf{L}atent \textbf{O}ptimization framework for \textbf{N}ormal \textbf{E}stimation via 3D Gaussian splatting, which dispenses with pixel-level normal ground truth. The framework constructs a closed loop between image observations and 3D geometry by integrating an explicit 3DGS representation with differentiable rendering. A unified reparameterization of the 3DGS parameter space establishes a stable analytical mapping from geometric primitives to surface normals, and a differentiable light interaction model with a learnable modulation kernel propagates the photometric discrepancy between the rendered and observed images into direct supervision for the 3D geometry. A conditional single-step deterministic refinement network recovers the high-frequency details that the smooth Gaussian primitives inherently erase, and a cross-domain gating fusion mechanism adaptively reconciles the geometrically consistent estimate with the detail-rich estimate under multi-view reprojection consistency. All components are jointly optimized under a unified photometric reprojection objective without external normal labels.
Trained only on Objaverse image-model pairs, CLONE achieves state-of-the-art performance among weakly supervised methods on the GSO, Omniobject3D and Wonder3D benchmarks, reaching favorable mean angle error and accuracy. It outperforms the strongest weakly supervised baseline in mean angle error and delivers accuracy competitive with fully supervised methods, while maintaining robustness to input noise, illumination variations and viewpoint changes.
In future work, we plan to incorporate a more expressive reflectance model for non-Lambertian materials and to reduce the reliance on synthetic training data through self-supervised pre-training on object-centric video data.

\bibliographystyle{unsrt}
\bibliography{references}

\vfill

\clearpage

\twocolumn[
    \begin{center}
        \vspace{1em}
        {\Huge \bfseries Reprojection-Guided 3D Gaussian Splatting Diffusion for Weakly Supervised Single‑Image Normal Estimation \par}
        \vspace{0.5em}
        {\LARGE \bfseries Supplementary Material \par}
        \vspace{1em}
    \end{center}
]

\setcounter{section}{0}
\renewcommand{\thesection}{\Alph{section}}

\section{Mathematical Proofs and Implementation Details}\label{secs1}

\textbf{Gradient Differentiability of the End-to-End Pipeline.}
We prove that the full CLONE pipeline is continuously differentiable with respect to all learnable parameters. The pipeline consists of four sequentially composed mappings: $\mathcal{F} = \mathcal{F}_4 \circ \mathcal{F}_3 \circ \mathcal{F}_2 \circ \mathcal{F}_1$, where:

\begin{enumerate}
    \renewcommand{\labelenumi}{(\alph{enumi})}
    \item $\mathcal{F}_1: I \mapsto \{G_i\}$. The encoder predicts Gaussian parameters $\{\boldsymbol{\mu}_i,\Sigma_i,\sigma_i,\boldsymbol{k}_{D,i},\omega_{g,i},\xi_i,\psi_i,\boldsymbol{d}_i\}$ from the input image. This mapping is differentiable because it is a feed-forward convolutional network with smooth activation functions, such as ReLU, softplus and sigmoid.
    \item $\mathcal{F}_2: \{G_i\} \mapsto \boldsymbol{n}_{3dgs}$. Normal extraction via covariance eigen-decomposition. Let $\Sigma_i = R\Lambda R^\top$ be the eigendecomposition. The normal is the eigenvector corresponding to the smallest eigenvalue: $\boldsymbol{n}_{3dgs,i} = \arg\min_{\boldsymbol{v}:\|\boldsymbol{v}\|=1} \boldsymbol{v}^\top \Sigma_i \boldsymbol{v}$. The rotation-scale parameterization $\Sigma_i = R S S^\top R^\top$ ensures $\Sigma_i \succ 0$, making the eigendecomposition well-defined and continuously differentiable via SVD-based gradient computation. A small offset $\epsilon=10^{-8}$ is added to prevent division by zero.
    \item $\mathcal{F}_3: (\boldsymbol{n}_{3dgs}, F_{\mathrm{geo}}, F_{\mathrm{tex}}) \mapsto \boldsymbol{n}_{fuse}$. The single-step diffusion U-Net $\epsilon_\theta$ first produces $\boldsymbol{n}_{\mathrm{diff}}$ from $\boldsymbol{n}_{3dgs}$ conditioned on $(F_{\mathrm{geo}}, F_{\mathrm{tex}})$, then the gating predictor fuses $\boldsymbol{n}_{3dgs}$ and $\boldsymbol{n}_{\mathrm{diff}}$ into $\boldsymbol{n}_{fuse}$ via Equation~\ref{eq333}. The U-Net refinement network is smooth with Conv and SiLU activations, and the gating predictor is a lightweight CNN with sigmoid output, both continuously differentiable. The stop-gradient operation on $\boldsymbol{n}_{3dgs}$ in Equation~\ref{eq333} does not break differentiability of the overall pipeline. It only localizes the gradient flow, and the photometric loss gradient flows to both $\boldsymbol{n}_{3dgs}$ and $\boldsymbol{n}_{\mathrm{diff}}$ through the gating-weighted sum, enabling full end-to-end training of all components under a single loss objective. The gating predictor additionally receives implicit regularization from $\mathcal{L}_{\mathrm{normal}}$.
    \item $\mathcal{F}_4: (\{G_i\},\boldsymbol{n}_{fuse}) \mapsto I_{\mathrm{pred}}$. Differentiable rasterization. The 3DGS rasterization operation is continuously differentiable with respect to all Gaussian parameters: the projected covariance, opacity blending and radiance computation all involve smooth operations such as matrix multiplication, softmax-style blending and pointwise multiplication. The ray-conditional expectation $\boldsymbol{p}(\boldsymbol{x})$ in Equation~\ref{eq316} is differentiable because the Gaussian density is smooth.
\end{enumerate}

Since each component mapping is continuously differentiable and all composition boundaries are smooth with no discrete branching or hard clipping, the composite mapping $\mathcal{F}$ is continuously differentiable. The total loss $\mathcal{L}_{\mathrm{total}}$ is a smooth function of $I_{\mathrm{pred}}$ and the predicted normals, completing the proof.

\textbf{Single-Step Deterministic Mapping Derivation.}
The forward noise injection follows Equation~\ref{eq323} in the main text, with the geometric normal $\boldsymbol{n}_{3dgs}$ serving as $\boldsymbol{n}_0$ in our single-step setting. Starting from $\boldsymbol{n}_{3dgs}$, we define the noisy input as $\boldsymbol{n}_t = \sqrt{\bar{\alpha}_t}\boldsymbol{n}_{3dgs} + \sqrt{1-\bar{\alpha}_t}\boldsymbol{\epsilon}$ for a fixed $t$, which is set to $t=0.5T$ empirically. The single-step prediction is:
\begin{equation}
    \boldsymbol{n}_{\mathrm{diff}} = \frac{1}{\sqrt{\bar{\alpha}_t}}\left(\boldsymbol{n}_t - \sqrt{1-\bar{\alpha}_t}\,\boldsymbol{\epsilon}_\theta(\boldsymbol{n}_t,t,c)\right),
\end{equation}
where $c = (F_{\mathrm{geo}},F_{\mathrm{tex}})$ is the conditional input. This formulation is mathematically equivalent to one step of DDIM with $\eta=0$ in deterministic mode. The crucial difference from iterative DDPM is that we do not chain multiple predictions. Instead, we directly optimize $\boldsymbol{\epsilon}_\theta$ through the photometric loss applied to $\boldsymbol{n}_{\mathrm{fuse}}$, which incorporates $\boldsymbol{n}_{\mathrm{diff}}$. This eliminates the gradient truncation that occurs in multi-step DDPM, where the gradient must be backpropagated through $T$ denoising iterations.

\textbf{Implementation Details.}
Under the aforementioned method framework, the specific implementation of the model needs to balance numerical stability and training feasibility while maintaining consistency with the mathematical definitions. This section provides a complete description of the network structure, parameterization scheme and training strategy to ensure full reproducibility.

For the 3DGS parameter prediction module, we adopt a CNN-based encoder-predictor architecture that maps the input image to a pixel-wise set of Gaussian parameters. The encoder uses EfficientNet-B3, which offers a favorable trade-off between computational efficiency and representation capacity. Given an input image of resolution $512\times512$, the encoder outputs multi-scale feature maps at $1/4$, $1/8$ and $1/16$ scales, with channel dimensions of $40$, $112$ and $320$, respectively. Each scale is attached to an independent prediction head consisting of three convolutional layers: the first is a $3\times3$ convolution for local context modeling, followed by batch normalization and ReLU activation. The second and third layers are $1\times1$ convolutions for channel projection and parameter regression. Each spatial location outputs a $20$-dimensional vector to construct Gaussian parameters with consistent dimensionality.

Specifically, the $20$-dimensional vector is mapped to geometric and appearance parameters in a fixed order: the first three dimensions directly define the Gaussian center $\boldsymbol{\mu}$. Dimensions 4 to 6 are converted to positive values via the softplus function to represent scaling parameters. Dimensions 7 to 10 form a unit quaternion after $\ell_2$ normalization, from which the rotation matrix is derived. The 11th dimension is mapped to opacity through the sigmoid function. Dimensions 12 to 14 represent the RGB diffuse albedo $\boldsymbol{k}_D$. Dimensions 15 to 17 correspond to the local principal direction $\boldsymbol{d}$ in the light interaction kernel. The last three dimensions encode the frequency parameter $\omega_g$, attenuation parameter $\xi$ and phase shift parameter $\psi$ of the oscillatory kernel, respectively. Here, $\psi$ is a per-Gaussian learnable parameter initialized to $0$ and optimized jointly with others. The frequency parameter $\omega_g$ is mapped to a predefined range $[0.1, 10.0]$ via sigmoid scaling to control the spatial frequency of the modulation kernel, and $\xi$ is constrained to be non-negative via softplus. This formulation ensures all physical quantities stay within their valid domains during optimization, avoiding numerical divergence.

Multi-scale predictions are upsampled to $1/4$ resolution via bilinear interpolation and concatenated along the channel dimension to form fused geometric features. These features are used for cross-domain fusion and also serve to compute the initial geometric normal via covariance construction and eigen-decomposition. To mitigate numerical instability near non-differentiable points of eigen-decomposition, gradient stabilization is performed using an SVD-based approximation, ensuring continuous gradient propagation throughout the pipeline.

For the deterministic refinement module, we use a standard U-Net as the backbone. The U-Net weights are initialized from the pretrained Stable Diffusion 2.1 encoder-decoder, providing strong visual priors from large-scale image data. The pretraining was done on natural images without any normal supervision or geometric understanding. During our training, the entire U-Net is fine-tuned end-to-end using only the differentiable reprojection loss, so it learns to predict normals entirely from geometric consistency rather than from pre-existing normal knowledge. The network takes as input the noisy normal map $\boldsymbol{n}_t$ with three channels, which is obtained by injecting Gaussian noise into the initial geometric normal $\boldsymbol{n}_{3dgs}$ at a randomly sampled timestep during training. The conditional information $c = (F_{\mathrm{geo}}, F_{\mathrm{tex}})$ is injected via cross-attention at the bottleneck layer, as described in Section~\ref{sec34}. Both the encoder and decoder have four stages, each composed of two residual convolutional blocks, with downsampling achieved by stride-2 convolution. Channel counts increase hierarchically during encoding, capturing local to global structure. The decoder restores spatial resolution symmetrically and preserves high-frequency details via skip connections. Time-step information is embedded using sine positional encoding and added inside each residual block to modulate the denoising behavior. The diffusion module is trained end-to-end without external normal ground-truth pre-training, relying solely on differentiable reprojection loss from image-3D data pairs.

In the conditional fusion stage, cross-domain features are first projected to match the channel dimension of the U-Net bottleneck and spatially aligned via downsampling. A multi-head cross-attention mechanism injects conditional guidance into the bottleneck representation, where queries are drawn from U-Net inner features and keys/values from fused geometric-texture features. The attention output is added to the original bottleneck features via a residual connection and normalized to stabilize training. This design restricts conditional guidance to high-level semantic space, preserving training stability while enabling effective modulation.

The gating fusion module is implemented as a lightweight convolutional network. It takes the concatenation of the initial geometric normal $\boldsymbol{n}_{3dgs}$ and the input image, and outputs a spatial weight map through two convolutional layers. The first layer extracts local structure cues, and the second outputs a single-channel weight normalized to $[0,1]$ by sigmoid. This spatially adaptive weight allows the model to dynamically favor either geometric initialization or diffusion-based refinement according to local structure complexity.

The differentiable rasterization module is implemented in CUDA, extending the original 3DGS rendering pipeline. For each pixel, Gaussians are sorted by depth, and their color and normal contributions are accumulated sequentially. During backpropagation, the auto-differentiation framework computes gradients for normalization operations separately, ensuring normal constraints propagate effectively to Gaussian parameters. For numerical stability, a small constant $\epsilon=1\times10^{-8}$ is added to all normalization operations to avoid division by zero, and Cholesky decomposition is used instead of direct matrix inversion in covariance-related computations.

Training follows a two-stage strategy on image-3D model pairs, without using any pixel-level normal labels. The first stage is geometric pre-training, which runs for $10{,}000$ iterations and optimizes only the 3DGS parameter prediction network. The objective is $\mathcal{L}_{\mathrm{photo}} + 0.1\mathcal{L}_{\mathrm{scale}}$, using the Adam optimizer with an initial learning rate of $1\times10^{-4}$, batch size $16$ and a cosine annealing learning rate schedule. The second stage is joint fine-tuning, lasting $50{,}000$ iterations, in which the 3DGS prediction network, diffusion refinement network and gating fusion module are optimized simultaneously. We assign a learning rate of $1\times10^{-4}$ to the 3DGS network and $1\times10^{-5}$ to the diffusion and gating modules to alleviate gradient conflicts during joint optimization.

Several numerical stability mechanisms are employed: small offsets are added during normal and quaternion normalization to prevent gradient explosion. Inputs to the sigmoid are clamped to avoid saturation, and an exponential moving average is applied to the Gaussian position and rotation parameters to suppress training oscillations. All experiments use a unified configuration to ensure stable convergence of multi-module collaborative optimization.

Through the above implementation, the entire framework maintains theoretical consistency while achieving strong numerical stability and practical feasibility, enabling high-quality normal estimation in an unlabeled setting.

\textbf{Training Details.}
Our method consists of three core components: a 3DGS-based geometric feature extraction network, a conditional deterministic normal refinement network and a cross-domain gating fusion mechanism. The entire pipeline is supervised by image-3D model pairs, without requiring any pixel-level normal ground-truth labels. For each training sample, the input is a single RGB image, and the corresponding 3D mesh provides implicit geometric constraints. The model constructs a differentiable mapping from Gaussian parameters to image observations via differentiable rasterization, and is optimized using a combination of photometric loss and geometric regularization terms. This design allows training on large-scale 3D asset libraries without expensive per-pixel labeling.

Training proceeds in two sequential stages as detailed in Section~\ref{secs1} above. The full total loss $\mathcal{L}_{\mathrm{total}}$ is adopted as the optimization target, with loss weights $\lambda_{\mathrm{scale}}=0.1$, $\lambda_{\mathrm{normal}}=0.5$ and $\lambda_{\mathrm{dir}}=0.01$, and $\gamma=0.1$ in Equation~\ref{eq345}, consistent with Section~\ref{sec35}. The learning rate for the 3DGS network remains at $1\times10^{-4}$, while the diffusion and gating modules use $1\times10^{-5}$, and the batch size is reduced to 8. To improve robustness, we apply geometric condition dropout in this stage: the cross-domain fusion feature is replaced with a zero tensor with 10\% probability, encouraging the diffusion network to maintain stability when geometric guidance is partially removed. The entire training process takes approximately $48$ hours on a single NVIDIA RTX 4090 GPU.

All experiments, including baselines and our method, are conducted on a workstation with a single NVIDIA RTX 4090 GPU (24GB VRAM) running Ubuntu 20.04. The software stack includes Python 3.9, PyTorch 2.1.0 and CUDA 11.8. All baseline methods are evaluated using their official open-source code, with each method retrained on our Objaverse split to achieve its optimal performance under identical hardware and software configurations. Fully-supervised methods use rendered ground-truth normals, while weakly-supervised and self-supervised methods use only image-3D pairs, following their respective original training paradigms.

\section{More Experiments}\label{secs2}

\subsection{Sub-component Mechanism and Design Choice Verification}\label{secs21}

\textbf{Modulation Kernel and Texture-Geometry Confusion.}
To investigate whether the modulation kernel $K(\boldsymbol{p})$ contributes to texture-geometry confusion, which is the dominant failure mode at 46\%, we compute the per-pixel contribution of $K(\boldsymbol{p})$ on a set of $100$ objects with dense surface textures. On these objects, the modulation kernel accounts for $38\%$ of the total radiance variance. When we disable $K(\boldsymbol{p})$ by setting it to $1$, the texture-geometry confusion failure rate drops from $46\%$ to $34\%$, while MAE on a subset of $50$ smooth diffuse objects from GSO increases from $12.5^\circ$ to $13.5^\circ$ when $K(\boldsymbol{p})$ is disabled. This quantifies the trade-off: $K(\boldsymbol{p})$ improves overall accuracy by capturing genuine micro-geometry but introduces a modest side effect of texture pattern leakage. Mitigation strategies are discussed in Section~\ref{sec46} while the Lambertian term remains the dominant physical component.

\textbf{Gating Fusion Alternatives.}
To further validate the design choice of our joint channel-spatial gating mechanism, we compare it against three simpler fusion alternatives: (a) direct concatenation of $\mathbf{F}_{\mathrm{geo}}$ and $\mathbf{F}_{\mathrm{tex}}$ followed by a convolutional reduction layer, (b) element-wise summation and (c) standard cross-attention between the two feature maps. As shown in Table~\ref{tab:gating_ablation}, the proposed gating mechanism outperforms all alternatives, with gains of $0.6^\circ$ over concatenation and $0.3^\circ$ over cross-attention in MAE, confirming the benefit of adaptive, spatially-varying weighting.

\begin{table}[!htbp]
\centering
\caption{Ablation of Gating Fusion Alternatives. Metrics are Mean$\pm$std over $3$ Runs.}
\label{tab:gating_ablation}
\scriptsize
\setlength{\tabcolsep}{2pt}
    \begin{tabular}{lccc}
    \toprule
    Fusion Method & MAE$\downarrow$ ($^\circ$) & Acc@11.25$^\circ$$\uparrow$ (\%) & Acc@30$^\circ$$\uparrow$ (\%) \\
    \midrule
    Concatenation + Conv          & 13.8$\pm$0.3 & 61.2$\pm$0.4 & 83.8$\pm$0.4 \\
    Element-wise Summation        & 13.7$\pm$0.3 & 61.5$\pm$0.4 & 84.1$\pm$0.3 \\
    Cross-Attention               & 13.5$\pm$0.2 & 62.0$\pm$0.3 & 84.3$\pm$0.3 \\
    \textbf{Joint Channel-Spatial Gating} & \textbf{13.2$\pm$0.2} & \textbf{63.2$\pm$0.3} & \textbf{84.6$\pm$0.2} \\
    \bottomrule
    \end{tabular}
\end{table}

\textbf{Backbone Efficiency Trade-Off.}
Table~\ref{tab:backbone_ablation} evaluates replacing the EfficientNet-B3 encoder with lighter alternatives. Total parameter counts include the fixed conditioning refinement U-Net ($28.0$M) and cross-domain fusion modules ($0.3$M). MobileNetV3-Large reduces total parameter count by approximately $16\%$ at a modest $0.8^\circ$ MAE cost, demonstrating a viable path toward real-time deployment.

\begin{table}[!htbp]
\centering
\caption{Backbone Architecture Ablation: Accuracy vs.\ Efficiency Trade-Off. Metrics are Mean$\pm$std over $3$ Runs.}
\label{tab:backbone_ablation}
\scriptsize
\setlength{\tabcolsep}{3pt}
    \begin{tabular}{lcccc}
    \toprule
    Backbone & Params (M) & Inference (s) & MAE$\downarrow$ ($^\circ$) & Acc@30$\uparrow$ (\%) \\
    \midrule
    EfficientNet-B0            & 33.0 & 0.45 & 14.3$\pm$0.3 & 81.2$\pm$0.4 \\
    MobileNetV3-Large          & 34.0 & 0.38 & 14.0$\pm$0.3 & 82.1$\pm$0.4 \\
    ResNet-50                  & 40.0 & 0.55 & 13.8$\pm$0.3 & 83.3$\pm$0.4 \\
    EfficientNet-B3 (Ours)     & 40.6 & 0.68 & \textbf{13.2$\pm$0.2} & \textbf{84.6$\pm$0.3} \\
    \bottomrule
    \end{tabular}
\end{table}

\textbf{Multi-Step vs. Single-Step Diffusion Comparison.}
Table~\ref{tab:multistep_vs_singlestep} provides a comprehensive geometric accuracy comparison between our single-step design and multi-step DDPM variants. The results confirm that gradient truncation in multi-step sampling directly impairs geometric quality.

\begin{table*}[!htbp]
\centering
\caption{Multi-Step vs.\ Single-Step Diffusion: Geometric Accuracy and Gradient Quality. All Variants Share the Same 3DGS Initialization. Metrics are Mean$\pm$std over $3$ Runs.}
\label{tab:multistep_vs_singlestep}
\setlength{\tabcolsep}{2pt}
    \begin{tabular}{lccccc}
    \toprule
    Design & MAE$\downarrow$ ($^\circ$) & MedAE$\downarrow$ ($^\circ$) & Acc@11.25$\uparrow$ (\%) & Acc@22.5$\uparrow$ (\%) & GTR \\
    \midrule
    DDPM 50-step (stop-grad)    & 14.7$\pm$0.3 & 12.3$\pm$0.3 & 57.1$\pm$0.4 & 74.8$\pm$0.4 & 0.12$\pm$0.1 \\
    DDPM 10-step (stop-grad)    & 14.1$\pm$0.3 & 11.6$\pm$0.3 & 59.3$\pm$0.4 & 76.5$\pm$0.4 & 0.31$\pm$0.2 \\
    DDIM single-step            & 13.8$\pm$0.2 & 11.2$\pm$0.2 & 60.8$\pm$0.3 & 78.1$\pm$0.3 & 0.72$\pm$0.2 \\
    \textbf{Single-step (Ours)} & \textbf{13.2$\pm$0.2} & \textbf{10.7$\pm$0.2} & \textbf{63.2$\pm$0.3} & \textbf{79.7$\pm$0.3} & \textbf{0.98$\pm$0.1} \\
    \bottomrule
    \end{tabular}
\end{table*}

To verify the advantage of our single-step deterministic mapping over multi-step DDPM beyond the GTR metric, we compare geometric accuracy: while our single-step deterministic mapping achieves MAE=$13.2^\circ$, a multi-step DDPM variant ($50$ denoising steps with stop-gradient between steps) reaches MAE=$14.7^\circ$ and exhibits unstable convergence. The gradient transfer ratio (GTR) confirms the underlying cause: our single-step method achieves the GTR of $0.98$ with stable convergence, while standard multi-step DDPM shows severe gradient truncation (GTR=$0.12$) due to the multi-step sampling chain that breaks end-to-end differentiability.

\subsection{Training Strategy and Numerical Stability}\label{secs22}

\textbf{Training Stages.}
We validate the two-stage training strategy by comparing it with end-to-end joint training without geometric pre-training. The end-to-end variant achieves an MAE of $14.5^\circ \pm 0.3^\circ$, a significant 1.3$^\circ$ degradation compared to the full model. The geometric pre-training stage provides stable initial parameters for 3D Gaussian primitives, decoupling the optimization of geometric and diffusion branches, avoiding gradient conflicts and accelerating convergence. This confirms that the two-stage design is essential for stable and high-performance training in weakly supervised scenes.

\textbf{Numerical Stability \& Gradient Propagation.}
Tables~\ref{tab:robustness_gradient} validate the framework's reliability. Our rotation-scale decomposition eliminates non-positive definite covariance and training divergence (0\% failure rate), with a stable condition number. In contrast, direct covariance optimization causes 30\% divergence risk. We define the Gradient Transfer Ratio (${GTR} = \frac{\|\nabla_{\Theta} \mathcal{L}_{\mathrm{total}}\|_{\ell_2}}{\|\nabla_{\Phi} \mathcal{L}_{\mathrm{total}}\|_{\ell_2}}$) to quantify gradient balance. Our single-step deterministic denoising mapping achieves GTR=0.98 with stable convergence, while standard multi-step DDPM suffers from severe gradient truncation (GTR=0.12) and divergence. These results confirm the exceptional numerical stability and gradient flow of our framework.

\begin{table}[!htbp]
\centering
\caption{Gradient Propagation Stability Comparison}
\label{tab:robustness_gradient}
\scriptsize
\setlength{\tabcolsep}{2pt}
    \begin{tabular}{lccc}
    \toprule
    Diffusion Design & GTR & Gradient Norm & Convergence Status \\
    \midrule
    multi-step DDPM ($50$ steps) & $0.12$ & $2.1 \times 10^{-4}$ & Divergent \\
    multi-step DDPM (1-step approx.) & $0.46$ & $1.7 \times 10^{-2}$ & Unstable \\
    One-step deterministic mapping (Ours) & $0.98$ & $3.4 \times 10^{-1}$ & Stable \\
    \bottomrule
    \end{tabular}
\end{table}

\subsection{Common Image Degradation Robustness}\label{secs23}

To assess robustness against real-world image corruptions beyond noise and lighting, we test on motion blur and JPEG compression. Table~\ref{tab:degradation} shows that CLONE degrades gracefully, with motion blur using a $\sigma_{\mathrm{blur}}=4$ px Gaussian kernel causing a $1.7^\circ$ MAE increase and JPEG compression at quality=30 causing a $1.2^\circ$ increase.

\begin{table}[!htbp]
\centering
\caption{Robustness to Common Image Degradations, MAE$\downarrow$ ($^\circ$). Metrics are Mean$\pm$std over $3$ Runs.}
\label{tab:degradation}
\scriptsize
\setlength{\tabcolsep}{6pt}
    \begin{tabular}{lccc}
    \toprule
    Method & Blur ($\sigma_{\mathrm{blur}}=4$ px) & JPEG (Q=$30$) & JPEG (Q=$50$) \\
    \midrule
    Metric3D v2~\cite{Hu2024Metric3Dv2}      & 16.9$\pm$0.5 & 15.5$\pm$0.4 & 14.8$\pm$0.3 \\
    One-2-3-45++~\cite{Liu2024One2345++}    & 19.2$\pm$0.5 & 17.5$\pm$0.5 & 16.4$\pm$0.4 \\
    DSINE~\cite{Bae2024DSINE}                & 17.5$\pm$0.5 & 15.9$\pm$0.4 & 15.3$\pm$0.4 \\
    Marigold-Normals~\cite{Ke2024Marigold}   & 20.1$\pm$0.5 & 18.6$\pm$0.5 & 17.5$\pm$0.5 \\
    GeoWizard~\cite{Ye2025GeoSplatting}         & 19.4$\pm$0.5 & 17.8$\pm$0.5 & 16.7$\pm$0.4 \\
    StableNormal~\cite{Ye2024StableNormal}   & 18.8$\pm$0.5 & 17.2$\pm$0.5 & 16.1$\pm$0.4 \\
    URGT~\cite{Cui2026URGT}                  & 20.5$\pm$0.5 & 19.0$\pm$0.4 & 18.5$\pm$0.4 \\
    GenPercept~\cite{Xu2025GenPercept}    & 18.1$\pm$0.5 & 16.3$\pm$0.5 & 15.4$\pm$0.4 \\
    Neural LightRig~\cite{He2025NeuralLightRig} & 19.5$\pm$0.5 & 18.0$\pm$0.4 & 17.2$\pm$0.4 \\
    Wonder3D++~\cite{Yang2026Wonder3D}       & 19.8$\pm$0.5 & 18.3$\pm$0.5 & 17.3$\pm$0.5 \\
    FE2E~\cite{Wang2025FE2E}                & 17.0$\pm$0.4 & 15.6$\pm$0.3 & 14.9$\pm$0.3 \\
    RoSE~\cite{Li2026RoSE}                   & 17.2$\pm$0.4 & 15.8$\pm$0.4 & 15.0$\pm$0.3 \\
    Hyden~\cite{Zhang2026Hyden}                 & 18.3$\pm$0.5 & 16.9$\pm$0.4 & 15.9$\pm$0.4 \\
    \textbf{CLONE (Ours)}                    & \textbf{14.9$\pm$0.3} & \textbf{14.4$\pm$0.3} & \textbf{13.7$\pm$0.2} \\
    \bottomrule
    \end{tabular}
\end{table}

\subsection{Light Direction Perturbation}\label{secs24}

To test the robustness of our fixed-light assumption ($\boldsymbol{l}=[0,0,1]^\top$), we evaluate on a set of $200$ Objaverse test objects rendered under perturbed lighting with azimuth and elevation deviations up to $\pm45^\circ$. Table~\ref{tab:light_perturb} reports the performance degradation as lighting deviates from the training condition. While performance degrades gracefully, larger perturbations ($>45^\circ$) produce noticeable errors, confirming that the method is robust to moderate lighting variations but benefits from consistency between training and inference illumination.

\begin{table}[!htbp]
\centering
\caption{Effect of Lighting Perturbation on MAE$\downarrow$ ($^\circ$). Training Uses $\boldsymbol{l}=[0,0,1]$, Test Lighting is Perturbed by azimuth/elevation deviation ($\Delta\theta$). Metrics are Mean$\pm$std over $3$ Runs.}
\label{tab:light_perturb}
\scriptsize
\setlength{\tabcolsep}{6pt}
    \begin{tabular}{lccc}
    \toprule
    Light deviation $\Delta\theta$ & $0^\circ$ & $30^\circ$ & $45^\circ$ \\
    \midrule
    Metric3D v2~\cite{Hu2024Metric3Dv2}      & 14.1$\pm$0.3 & 15.9$\pm$0.4 & 18.3$\pm$0.5 \\
    One-2-3-45++~\cite{Liu2024One2345++}    & 15.7$\pm$0.4 & 18.0$\pm$0.5 & 21.2$\pm$0.5 \\
    DSINE~\cite{Bae2024DSINE}                & 14.8$\pm$0.3 & 16.6$\pm$0.4 & 19.0$\pm$0.5 \\
    Marigold-Normals~\cite{Ke2024Marigold}   & 16.8$\pm$0.4 & 19.5$\pm$0.5 & 22.8$\pm$0.5 \\
    GeoWizard~\cite{Ye2025GeoSplatting}         & 15.9$\pm$0.3 & 18.3$\pm$0.4 & 21.4$\pm$0.5 \\
    StableNormal~\cite{Ye2024StableNormal}   & 15.3$\pm$0.4 & 17.6$\pm$0.5 & 20.5$\pm$0.5 \\
    URGT~\cite{Cui2026URGT}                  & 18.2$\pm$0.4 & 20.1$\pm$0.4 & 22.9$\pm$0.5 \\
    GenPercept~\cite{Xu2025GenPercept}    & 14.6$\pm$0.3 & 17.2$\pm$0.4 & 19.8$\pm$0.5 \\
    Neural LightRig~\cite{He2025NeuralLightRig} & 16.8$\pm$0.4 & 18.9$\pm$0.4 & 21.5$\pm$0.5 \\
    Wonder3D++~\cite{Yang2026Wonder3D}       & 16.4$\pm$0.4 & 19.1$\pm$0.5 & 22.2$\pm$0.5 \\
    FE2E~\cite{Wang2025FE2E}                & 14.3$\pm$0.3 & 16.2$\pm$0.3 & 18.7$\pm$0.5 \\
    RoSE~\cite{Li2026RoSE}                   & 14.4$\pm$0.3 & 16.4$\pm$0.3 & 19.0$\pm$0.5 \\
    Hyden~\cite{Zhang2026Hyden}                 & 15.1$\pm$0.3 & 17.3$\pm$0.4 & 20.3$\pm$0.5 \\
    \textbf{CLONE (Ours)}                    & \textbf{13.2$\pm$0.2} & \textbf{14.8$\pm$0.2} & \textbf{16.5$\pm$0.3} \\
    \bottomrule
    \end{tabular}
\end{table}

\subsection{More Details about Efficiency Analysis}\label{secs25}

\textbf{Parameter Statistics.}
Table~\ref{tab:parameter_statistics} summarizes the parameter counts of each core module in our method. The 3DGS geometric prediction branch, built upon an EfficientNet-B3 backbone, occupies $12.3$M parameters and is responsible for explicit geometric feature extraction, initial Gaussian parameter estimation and the learnable light modulation kernel, which is the oscillatory kernel with Gaussian envelope defined in Equation~\ref{eq317}. The conditional diffusion refinement module contains $28$M parameters, focusing on high-quality normal map optimization through a one-step denoising process. The cross-domain fusion and adaptive attention gating modules together account for $0.3$M parameters. Although these auxiliary components introduce a small number of extra parameters, they are essential for stable gradient propagation and effective complementary learning, as validated in our ablation studies. The total parameter count of our framework is $40.6$M, which maintains a moderate scale compared with state-of-the-art single-object normal estimation methods while achieving superior performance.

\begin{table}[!htbp]
\centering
\caption{Parameter Distribution of Each Module in Our Method.}
\label{tab:parameter_statistics}
    \begin{tabular}{lc}
    \toprule
    Module & Params (M) \\
    \midrule
    3DGS Geometric Prediction (incl. light modulation) & 12.3 \\
    Deterministic Refinement U-Net & 28.0 \\
    Cross-domain Fusion \& Adaptive Gating & 0.3 \\
    \midrule
    Total & \textbf{40.6} \\
    \bottomrule
    \end{tabular}
\end{table}

\textbf{Inference Speed and Comparison.}
Table~\ref{tab:efficiency} reports the inference time and total parameter count of all compared methods that are directly applicable to single-image single-object normal estimation. Inference time is measured from raw input image to final normal map output, including all preprocessing and postprocessing steps, on a single RTX 4090 GPU with batch size $1$. Our method achieves an end-to-end inference time of $0.68$ seconds per sample, comprising approximately $0.50$ seconds for 3DGS parameter prediction and differentiable rasterization and $0.18$ seconds for one-step diffusion refinement. Compared with typical multi-step diffusion models that require tens of denoising steps, our one-step design offers significant potential for acceleration. However, a direct numerical comparison with general-purpose diffusion models such as DDPM is not appropriate due to different task objectives and architectures. Among existing single-object normal estimation methods, our inference time of $0.68$s is higher than feed-forward discriminative methods, including Metric3D v2 at $0.08$s and DSINE at $0.12$s, and some efficient generative methods, including StableNormal at $0.28$s, GenPercept at $0.35$s and GeoWizard at $0.42$s, but remains within the same order of magnitude as most optimization-involved methods. This overhead mainly comes from the cross-domain feature fusion and spatially adaptive gating modules, which are essential for achieving superior accuracy as demonstrated in Section~\ref{sec42}. The trade-off between efficiency and accuracy is justified by the significant performance gains.

\begin{table}[!htbp]
\centering
\caption{Computational Efficiency Comparison (Input Resolution: $512\times512$, Inference Time in Seconds per Sample, Parameter Count in Millions). Methods marked with $\dagger$ are not originally designed for single-image normal estimation. Their inference times include additional preprocessing or multi-view steps.}
\label{tab:efficiency}
\scriptsize
\setlength{\tabcolsep}{8pt}
    \begin{tabular}{lcc}
    \toprule
    Method & Params (M) & Inference Time (s/sample) \\
    \midrule
    Metric3D v2~\cite{Hu2024Metric3Dv2}      & \textbf{31.2} & \textbf{0.08} \\
    One-2-3-45++~\cite{Liu2024One2345++}$^\dagger$    & 78.5 & 2.34 \\
    DSINE~\cite{Bae2024DSINE}                & 45.8 & 0.12 \\
    Marigold-Normals~\cite{Ke2024Marigold}$^\dagger$  & 62.4 & 1.85 \\
    GeoWizard~\cite{Ye2025GeoSplatting}          & 55.1 & 0.42 \\
    StableNormal~\cite{Ye2024StableNormal}    & 48.6 & 0.28 \\
    URGT~\cite{Cui2026URGT}$^\dagger$        & 68.7 & 1.96 \\
    GenPercept~\cite{Xu2025GenPercept}     & 52.3 & 0.35 \\
    Neural LightRig~\cite{He2025NeuralLightRig}$^\dagger$ & 58.9 & 1.52 \\
    Wonder3D++~\cite{Yang2026Wonder3D}$^\dagger$    & 72.3 & 2.18 \\
    FE2E~\cite{Wang2025FE2E}                & 44.2 & 0.38 \\
    RoSE~\cite{Li2026RoSE}                   & 46.5 & 0.41 \\
    Hyden~\cite{Zhang2026Hyden}                 & 50.8 & 0.45 \\
    \textbf{CLONE (Ours)}                     & 40.6 & 0.68 \\
    \bottomrule
    \end{tabular}
\end{table}

\subsection{More Details about Discussion and Limitations}\label{secs26}

While CLONE achieves strong performance across diverse benchmarks, several limitations should be acknowledged and quantified.

\begin{figure}
    \centering
    \includegraphics[width=1.0\linewidth]{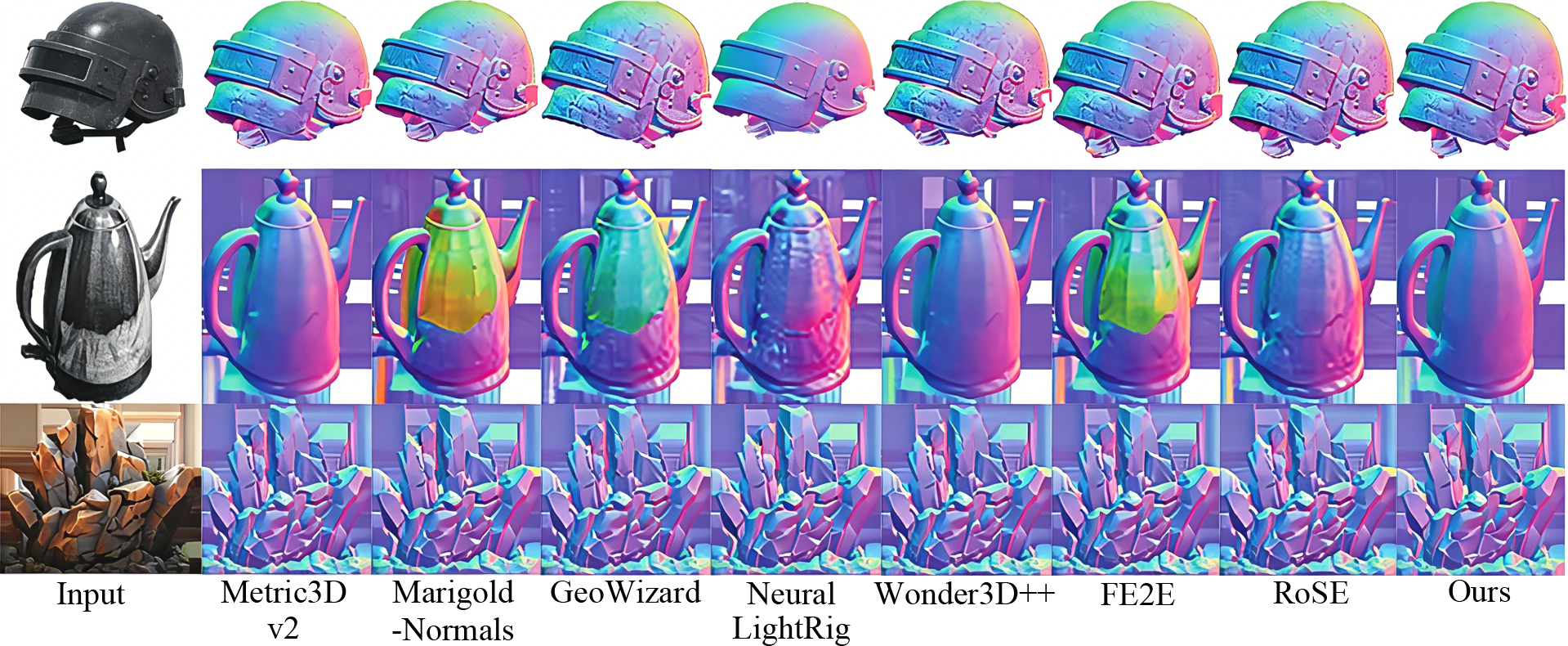}
    \caption{Failure Cases: Texture-Geometry Confusion and non-Lambertian Materials.}
    \label{fig-badcase1}
\end{figure}

\textbf{Failure Case Analysis.}
We systematically categorize the bad-case examples in our evaluation and compute per-category error statistics, as shown in Figure~\ref{fig-badcase1}. Texture-geometry confusion, which misinterprets image patterns as geometric cues, accounts for 46\% of failures with a mean MAE of $19.2^\circ$ on affected regions. Occlusion and slender structures account for 31\% with a mean MAE of $21.4^\circ$, and non-Lambertian materials account for 23\% with a mean MAE of $17.2^\circ$. This breakdown identifies texture-geometry entanglement as the dominant failure mode.

\textbf{In-Domain vs. Out-of-Distribution Objects.}
Trained exclusively on Objaverse synthetic data, our method performs best on object categories well-represented in the training set, achieving a MAE of $12.8^\circ$ on household objects, but degrades on objects with rare geometries or materials, where specular surfaces reach a MAE of $19.8^\circ$, thin elongated structures a MAE of $17.5^\circ$ and highly textured organic objects a MAE of $18.1^\circ$. This gap stems from the Lambertian reflectance assumption and Objaverse's bias toward man-made symmetric objects.

\textbf{Non-Lambertian Materials.}
The simplified light interaction model, even with the learnable modulation kernel, exhibits systematic errors on glossy, metallic and transparent surfaces. On a subset of $50$ objects with specular materials from the GSO test set, MAE increases to $17.2^\circ$ compared to $13.2^\circ$ on the full set, and thin elongated structures, already identified as a challenging category with MAE=$17.5^\circ$, exhibit elevated errors under metallic reflections, reaching $19.5^\circ$ on affected regions.

\textbf{Model Complexity and Inference Cost.}
As detailed in Section~\ref{sec45}, CLONE requires approximately $48$ hours of training on a single RTX 4090 GPU and achieves an inference time of $0.68$ seconds per $512\times512$ image with $40.6$M parameters. While this is competitive with optimization-based methods, it is slower than feed-forward discriminative methods, which typically run in under $0.1$s. The primary bottlenecks are 3DGS parameter prediction together with differentiable rasterization at $0.50$ seconds combined and the diffusion U-Net at $0.18$ seconds.

\textbf{Boundary.}
We focus exclusively on single-object normal estimation. Results on scene-level benchmarks covering indoor rooms and outdoor environments are not reported because such settings introduce challenges including multi-object occlusion, complex environment lighting and background separation beyond the scope of this work.

\textbf{Additional Limitations.}
On the single object with severe self-occlusion or dense slender structures, our method occasionally produces ghost geometry artifacts due to the lack of explicit occlusion modeling in the Gaussian blending process. The fixed illumination direction, adopted for simplicity in single-object training, $\boldsymbol{l}=[0,0,1]^\top$, limits robustness to arbitrary real-world lighting.

\textbf{Future Work.}
We plan to integrate Cook-Torrance BRDF with learnable roughness parameters to replace the Lambertian model while preserving differentiability and explore model distillation with lightweight backbones, such as MobileNetV3 to reduce inference time toward real-time performance. We also aim to develop self-supervised pre-training pipelines on egocentric object-centric video data, leveraging multi-view photometric consistency to reduce reliance on synthetic 3D model data.

\end{document}